\documentclass[10pt,twocolumn,letterpaper]{article}
\usepackage{authblk}
\PassOptionsToPackage{pagenumbers}{iccv}

\usepackage{iccv}         



\usepackage[accsupp]{axessibility}
\usepackage{amssymb}
\usepackage{pifont}
\usepackage{multirow}
\usepackage{commath}
\usepackage{makecell}
\usepackage{pifont}
\usepackage[table]{xcolor}
\usepackage{appendix}
\definecolor{iccvblue}{rgb}{0.21,0.49,0.74}
\usepackage[pagebackref,breaklinks,colorlinks,allcolors=iccvblue]{hyperref}
\newcommand{\cmark}{\ding{51}}
\newcommand{\xmark}{\ding{55}}

\usepackage{xr-hyper}

\makeatletter
\newcommand*{\addFileDependency}[1]{
  \typeout{(#1)}
  \@addtofilelist{#1}
  \IfFileExists{#1}{}{\typeout{No file #1.}}
}
\newcommand*{\newbibstartnumber}[1]{%
  \apptocmd{\thebibliography}{%
    \global\c@NAT@ctr #1\relax
    \addtocounter{NAT@ctr}{-1}%
  }{}{}%
}
\makeatother

\title{O-MaMa: Learning Object Mask Matching \\between Egocentric and Exocentric Views}

\author[]{Lorenzo Mur-Labadia\textsuperscript{*}}
\author[]{Maria Santos-Villafranca\textsuperscript{*}}
\author[]{Jesus Bermudez-Cameo}
\author[]{\\Alejandro Perez-Yus}
\author[]{Ruben Martinez-Cantin}
\author[]{}
\author[]{Jose J. Guerrero}

\affil[]{I3A - University of Zaragoza}

\begin{document}
\maketitle

{
  \renewcommand{\thefootnote}{\fnsymbol{footnote}}
  \footnotetext[1]{Equal contribution.}
  \renewcommand{\thefootnote}{} 
  \footnotetext{\begin{minipage}[t]{\columnwidth} Corresponding authors: \texttt{lmur@unizar.es},\\ \texttt{m.santos@unizar.es} \end{minipage}\\}
  \footnotetext{Code: \url{https://github.com/Maria-SanVil/O-MaMa}}
  \renewcommand{\thefootnote}{\arabic{footnote}} 
}

\begin{abstract}

Understanding the world from multiple perspectives is essential for intelligent systems operating together, where segmenting common objects across different views remains an open problem.
We introduce a new approach that re-defines cross-image segmentation by treating it as a mask matching task.
Our method consists of: 
(1) A Mask-Context Encoder that pools dense DINOv2 semantic features to obtain discriminative object-level representations from FastSAM mask candidates, 
(2) an Ego$\leftrightarrow$Exo Cross-Attention that fuses multi-perspective observations, 
(3) a Mask Matching contrastive loss that aligns cross-view features in a shared latent space, and
(4) a Hard Negative Adjacent Mining strategy to encourage the model to better differentiate between nearby objects.
O-MaMa achieves the state of the art in the Ego-Exo4D Correspondences benchmark, obtaining relative gains of +22 $\%$ and +76 $\%$ in the Ego2Exo and Exo2Ego IoU against the official challenge baselines, and a +13 $\%$ and +6 $\%$ compared with the SOTA with 1 $\%$ of the training parameters.

\end{abstract}    
\section{Introduction}
\label{sec:intro}

Nowadays, intelligent agents need to collaborate while performing cooperative tasks. This includes applications like multi-robot manipulation \cite{schmuck2019ccm,karrer2018collaborative}, augmented reality assistants \cite{vyas2017augmented,tran2023wearable}, and human-robot collaboration \cite{chang2021unfair,claure2022fairness}.
Perception plays a crucial role in this scenario, but each agent typically has access only to its own sensors or cameras, and each one perceives the environment from a different perspective. 
Consequently, understanding object correspondences between egocentric (first-person) and exocentric (third-person) views is essential to align multi-agent perception and establish a shared basis for interaction.
Despite the advances in segmentation \cite{long2015fully, xie2021segformer, liang2023open, minderer2023scaling} and object detection \cite{he2017mask, bolya2019yolact} from single images, cross-view segmentation between egocentric and exocentric perspectives remains an open challenge.

\begin{figure}[t]
    \centering
    \includegraphics[width=0.95\linewidth]{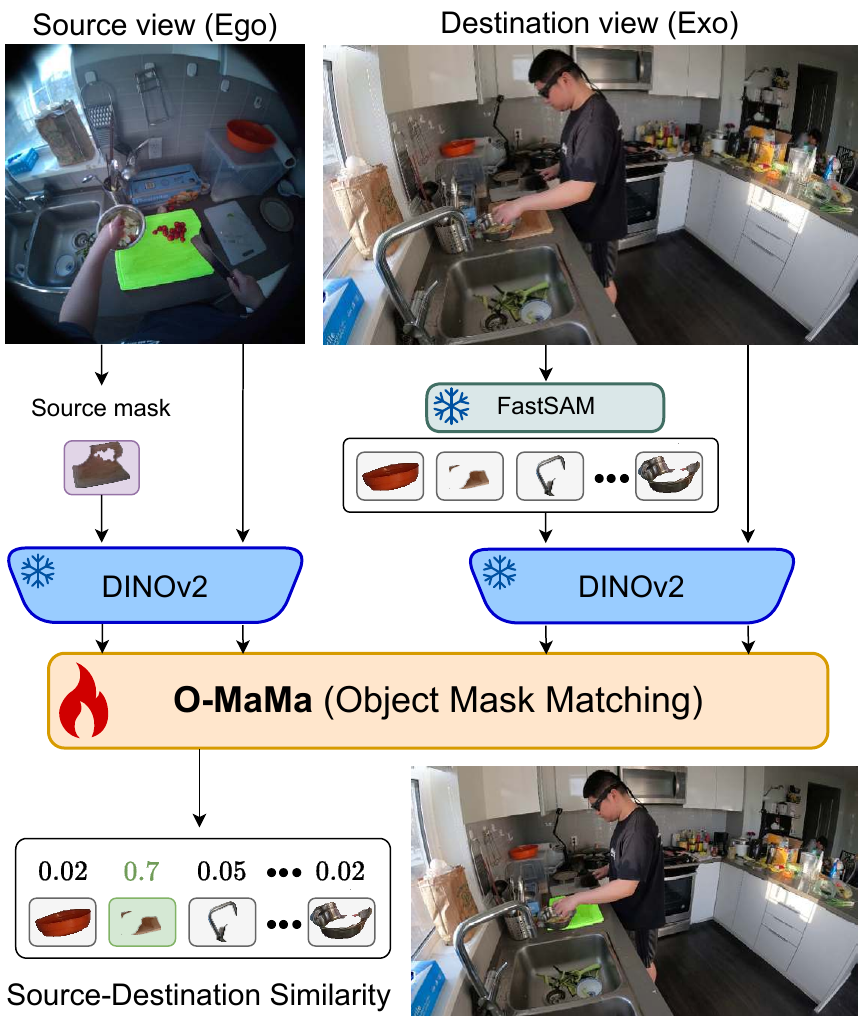}
    \caption{\textbf{Overview of our proposed Object Mask Matching (O-MaMa).} Instead of attempting the complex cross-view segmentation task, we obtain a set of mask candidates in the destination view using FastSAM. Through contrastive learning, we select the mask candidate that best matches the source mask.}
    \label{fig:teaser}
\end{figure}

In this paper, we address the Ego-Exo Correspondences task, where the goal is to predict an object's mask in one perspective given a query mask from the other.
Unlike traditional segmentation problems, this task introduces additional challenges, including drastic viewpoint transformations, scale variations, occlusions, and domain shifts due to differences in camera optics and imaging conditions.
While the exocentric view captures both the full environment and the person's body, it contains objects at multiple scales.
Conversely, the egocentric view offers rich details on hand-object interactions, but it is highly dynamic, suffering from motion blur and frequent occlusions due to the ongoing interactions.  
These challenges make establishing precise object-level correspondences across views particularly difficult, requiring fine-grained segmentation and strong cross-view semantic reasoning.

We propose simplifying the complex cross-image segmentation task by reformulating it as Object Mask Matching (O-MaMa) across ego and exo views, leveraging the excellent zero-shot segmentation capabilities of Segment Anything Models (SAM) \cite{kirillov2023segment}. An overview of our method is shown in \cref{fig:teaser}.
First, we extract a set of object mask candidates in the destination view using FastSAM \cite{zhao2023fast}. 
To obtain mask descriptors, we encode each object mask with a \textit{Mask-Context Encoder}, which pools dense DINOv2 \cite{oquab2023dinov2} semantic features from both object masks and their extended bounding boxes, combining discriminative object features with contextual information.
In cluttered scenes, nearby objects often share similar context while containing distinct object descriptors.
Therefore, we introduce a \textit{Hard Negative Adjacent Mining} strategy that selects neighboring object candidates to encourage the model to better differentiate between nearby objects.
Next, we incorporate cross-view global features using a novel \textit{Ego$\leftrightarrow$Exo Cross Attention} mechanism. 
Finally, we train the model with a \textit{Mask Matching Contrastive Loss}, which selects the best mask candidate in the destination view by learning a cross-view feature alignment that captures both global scene context and fine-grained object features.

Our method is both simple and effective, achieving state of the art in the Ego-Exo4D Correspondences benchmark \cite{grauman2024ego}. 
We obtain 45.8 (Ego2Exo) and 48.6 (Exo2Ego) IoU in the test v2 split, which represents a relative gain of +22.1 $\%$ and +76.4$ \%$ against the official challenge baselines \cite{grauman2024ego}, respectively.
Moreover, in the validation v1 set, O-MaMa scores 50.1 and 54.2 IoU in the Ego2Exo and Exo2Ego scenarios, showing a +13.1 $\%$ and +6.5 $\%$ against the previous SOTA model \cite{fu2024objectrelator} using only 1 $\%$ of training parameters.

\section{Related Works}
\label{sec:related}

\paragraph{Ego-Exo understanding.}
Exocentric (third-person) vision has been extensively studied in action recognition~\citep{hutchinson2021video, li2024videomamba}, segmentation~\citep{tsai2016video, kim2020video, woo2021learning} and tracking \cite{wang2024visual, ciaparrone2020deep}. 
Alternatively, egocentric (first-person) vision offers a unique viewpoint for capturing human behavior. The arrival of large-scale datasets \cite{grauman2022ego4d, damen2018scaling}
has driven progress in action recognition \cite{bansal2022my,radevski2023multimodal}, action anticipation \cite{mur2024aff, furnari2020rolling}, affordance segmentation \cite{mur2023multi, nagarajan2020ego}, and episodic memory \cite{mai2023egoloc, barmann2022did}. 
Since egocentric and exocentric views provide complementary visual cues of the scene, combining both perspectives has emerged as a promising direction for learning generalizable view-invariant representations.
Some works \cite{li2021ego, dou2024unlocking} improve model training in one perspective by leveraging data properties from the other view.
Other works explore the benefits of learning cross-view invariant features for action recognition \cite{yu2019joint, xue2023learning, rahmani2016knowledge}, 
affordance segmentation \cite{li2023locate}, activity progression \cite{donahue2024learning}, and temporal action segmentation \cite{quattrocchi2024synchronization}.
In contrast, we propose learning view invariant features at object level to match masks across synchronized views, which requires fine-grained pixel-level predictions.

\begin{figure*}[t]
    \centering
    \includegraphics[width=0.95\linewidth]{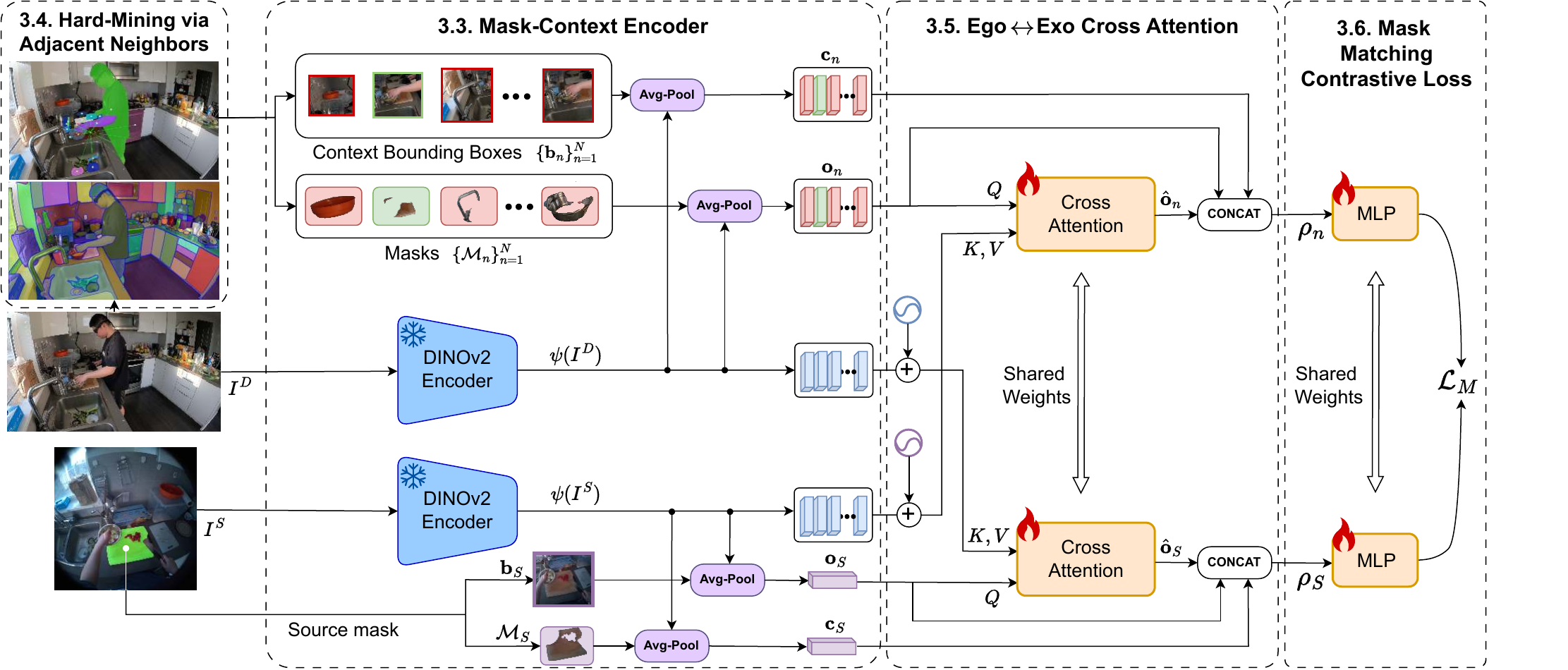}
    \caption{\textbf{O-MaMa architecture.} In the destination view, we generate a set of mask candidates with FastSAM. 
    We extract descriptors on both source and destination masks by pooling dense DINOv2 features, and we aggregate global cross-view features with respective cross-attention mechanisms. 
    We learn view-invariant features in a latent space via contrastive learning, and we select the most similar mask embedding to obtain the corresponding mask.}
    \label{fig:architecture}
\end{figure*}

\vspace{-1\baselineskip}
\paragraph{Learning correspondences.}
Traditionally, detector-based local feature matching methods first detect key-points in an image and then extract descriptors to establish correspondences following hand-crafted \cite{lowe2004distinctive, rublee2011orb, bay2008speeded} or learning \cite{detone2018superpoint, yi2016lift} approaches, using nearest neighbor search or an attentional graph neural network \cite{sarlin2020superglue} to find matches between the extracted interest points. 
More recently, detector-free methods \cite{sun2021loftr, jiang2021cotr, edstedt2024roma} have gained attention for their ability to directly obtain dense feature matches without an explicit keypoint detection. 
For instance, RoMa \cite{edstedt2024roma} achieves robust feature matching under extreme changes in scale, viewpoint and illumination.
While these methods establish correspondences between images of the same scene or object instances, semantic correspondence approaches \cite{liu2020semantic, cho2021cats, zhang2023tale} find dense matches between semantically similar images.
For example, Zhang \etal \cite{zhang2023tale} fuses DINOv2, that provides sparse but accurate matches, with stable diffusion features, which contain high-quality spatial information.
Here, we propose leveraging powerful semantic features from DINOv2 as well, but in a different context. Instead of keypoints or semantic parts, we establish correspondences between entire object masks across different perspectives.

\vspace{-1\baselineskip}
\noindent
\paragraph{Segmentation Models.}
Traditional approaches are categorized into semantic segmentation \cite{long2015fully, xie2021segformer}, instance segmentation \cite{he2017mask, bolya2019yolact, mur2023bayesian},
panoptic segmentation \cite{kirillov2019panoptic, cheng2020panoptic, mohan2021efficientps, cheng2022masked}, and video object segmentation \cite{perazzi2016benchmark, yang2019anchor, tokmakov2017learning}. 
SAM \cite{kirillov2023segment} introduced the promptable segmentation task \cite{li2024omg, ke2023segment, zhao2023fast, xiong2024efficientsam}, where fine-grained segmentation masks are generated from a spatial prompt (\ie, a point, or a bounding box).
Recently, several works \cite{lai2024lisa, zhang2024psalm, zou2023generalized} have leveraged the input token flexibility of Large Language Models (LLMs) to support diverse input conditions and output formats.

Despite the significant advances in single-image segmentation, few works address object segmentation across multiple images.
For instance, semi-supervised video object segmentation \cite{perazzi2016benchmark} methods segment an object along the video given its mask in the initial frame \cite{goyal2023tam, Oh_2019_ICCV, li2023unified, caelles20192019, xu2018youtube}.
Similarly, image co-segmentation models aim to identify common objects across different images \cite{rubinstein2013unsupervised, taniai2016joint, shen2022learning}.
Adapting LLMs to the cross-view segmentation, Object-Relator \cite{fu2024objectrelator} fine-tuned PSALM \cite{zhang2024psalm} with a dedicated module that enforces view-invariant embeddings through a self-supervised alignment.
In this work, we propose to reformulate the cross-view segmentation as a mask matching task, incorporating FastSAM \cite{zhao2023fast} in our pipeline to leverage its fine-grained, zero-shot segmentation capabilities; showing that we can achieve state-of-the-art results while significantly reducing the number of trainable parameters.

\section{Methods}
\label{sec:methods}

\subsection{Task Formulation}

Given a pair of images from two different views (egocentric and exocentric) and a query object mask $\mathcal{M}_{S}$ in the source view $I^S$, the objective of the Ego-Exo Correspondences task is to predict the corresponding object mask in the destination view $I^D$. 
In the Ego2Exo task, the source view corresponds to the egocentric image, and the destination view is the exocentric image, and the opposite happens in the Exo2Ego setting.
The task is restricted to use only visual information as input for the model, excluding semantic labels, object names, or camera pose information.

\begin{figure}[t]
    \centering
    \includegraphics[width=\linewidth]{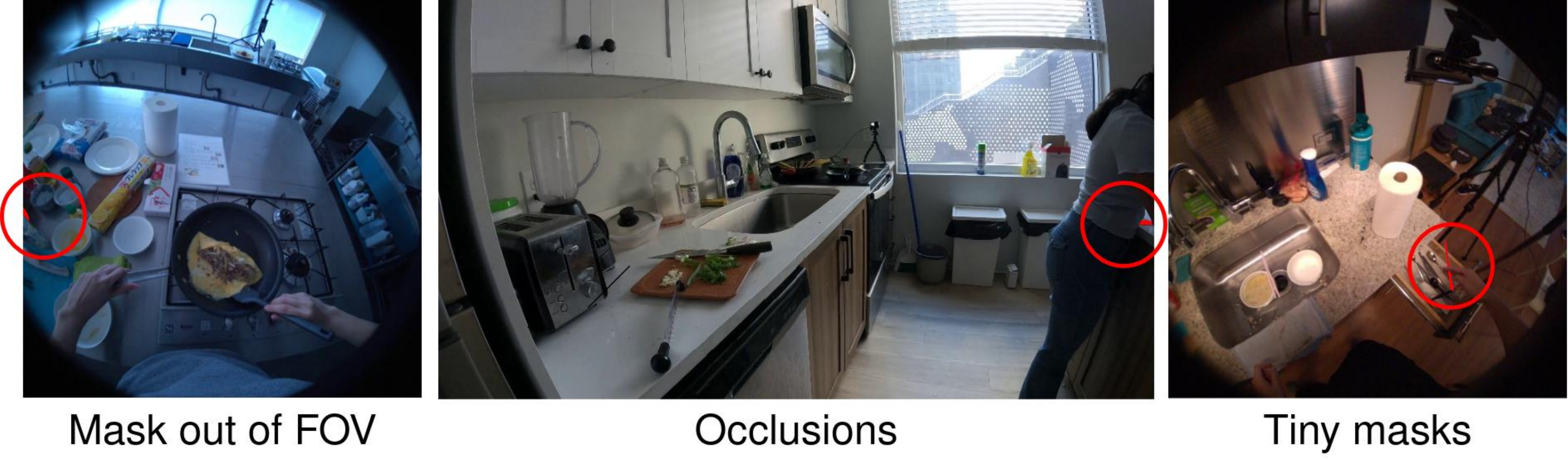}
    \caption{\textbf{Examples of complex scenarios.} The object may appear on the edges, be partially occluded or be extremely small.}
    \label{fig:complex_scenarios}
\end{figure}

Additionally, the task involves significant viewpoint variation and challenges due to the inherent characteristics of each viewpoint:
egocentric views suffer camera motion and occlusions due to the ongoing interactions,
while the exocentric perspective contains objects at multiple scales distributed along all the scene. See \cref{fig:complex_scenarios} for some examples of these limitations in the dataset.
All these factors require combining both a very fine-grained segmentation capability and a global cross-view understanding to effectively locate the corresponding objects across views.

\subsection{Method overview}

We reformulate the challenging cross-view segmentation task as a cross-view Object Mask Matching (O-MaMa) problem. This novel approach exploits the high-quality zero-shot segmentation capabilities exhibited by SAM \cite{kirillov2023segment, zhao2023fast} to simplify the segmentation problem.
\cref{fig:architecture} shows the architecture of our method.
The goal of O-MaMa is to select the object mask candidate in the destination view that matches the source mask best.
We first extract a set of object mask proposals in the destination view using FastSAM~\cite{zhao2023fast}, from which we compute object-level and contextual descriptors with the \textit{Mask-Context Encoder} (Section \ref{sec:mask_enc}). We also include cross-view global features through a novel \textit{Ego$\leftrightarrow$Exo Cross Attention} mechanism (Section \ref{sec:cross_attn}). 
We adopt a novel \textit{Mask Matching Contrastive Loss} (Section \ref{sec:loss}) to learn view-invariant features by minimizing the distance between paired samples and maximizing the distance between negative (unpaired) samples in a shared latent space. 
During training, we enforce the model to learn more robust and discriminative object descriptors with a \textit{Hard-Mining via Adjacent Neighbors} (Section \ref{sec:mining}).

\subsection{Mask-Context Encoder}
\label{sec:mask_enc}

We first generate a set of dense mask proposals in the destination view using FastSAM \cite{zhao2023fast}, which segments intuitive regions of the scene, such as entire objects, object parts or surfaces, with comparable quality to SAM \cite{kirillov2023segment}, while being 50$\times$ faster.
Specifically, from the destination image, $I^{D}$, we generate $N$ mask candidates $ \{ \mathcal{M}_{n} \}^{N}_{n = 1 }$.

Then, we compute a descriptor of each mask segment. 
We leverage DINOv2 \cite{oquab2023dinov2}, a self-supervised learning model, due to its high-level semantics, object decomposition capabilities and dense feature localization properties.
We extract a local object descriptor $\mathbf{o}_{n} = \text{Avg-Pool}(\mathcal{M}_{n}, \psi(I^{D}))$ by pooling the corresponding object mask from the DINOv2 feature map of the destination image, denoted as $\psi(I^D)$.
Some studies suggest \cite{garg2024revisit, bar2004visual, intraub1997representation} that humans leverage visual contextual associations among objects to represent scenes. Inspired by this, we extract a context descriptor $\mathbf{c}_{n} = \text{Avg-Pool}(\mathbf{b}_{n}, \psi(I^{D}))$ by pooling features from an extended bounding box $\mathbf{B}$ around each mask. 
In both cases, we upsample $\times4$ the DINOv2 feature map size in order to retain feature's regions fine granularity \cite{shlapentokh2024region}.
Similarly, we extract object $\mathbf{o}_{{S}}$ and context embeddings $\mathbf{c}_{{S}}$ of the source mask $\mathcal{M}_S$ in the other view.

\subsection{Hard-Mining via Adjacent Neighbors}
\label{sec:mining}

\begin{figure} [t]
    \centering
    \includegraphics[width=0.345\linewidth]{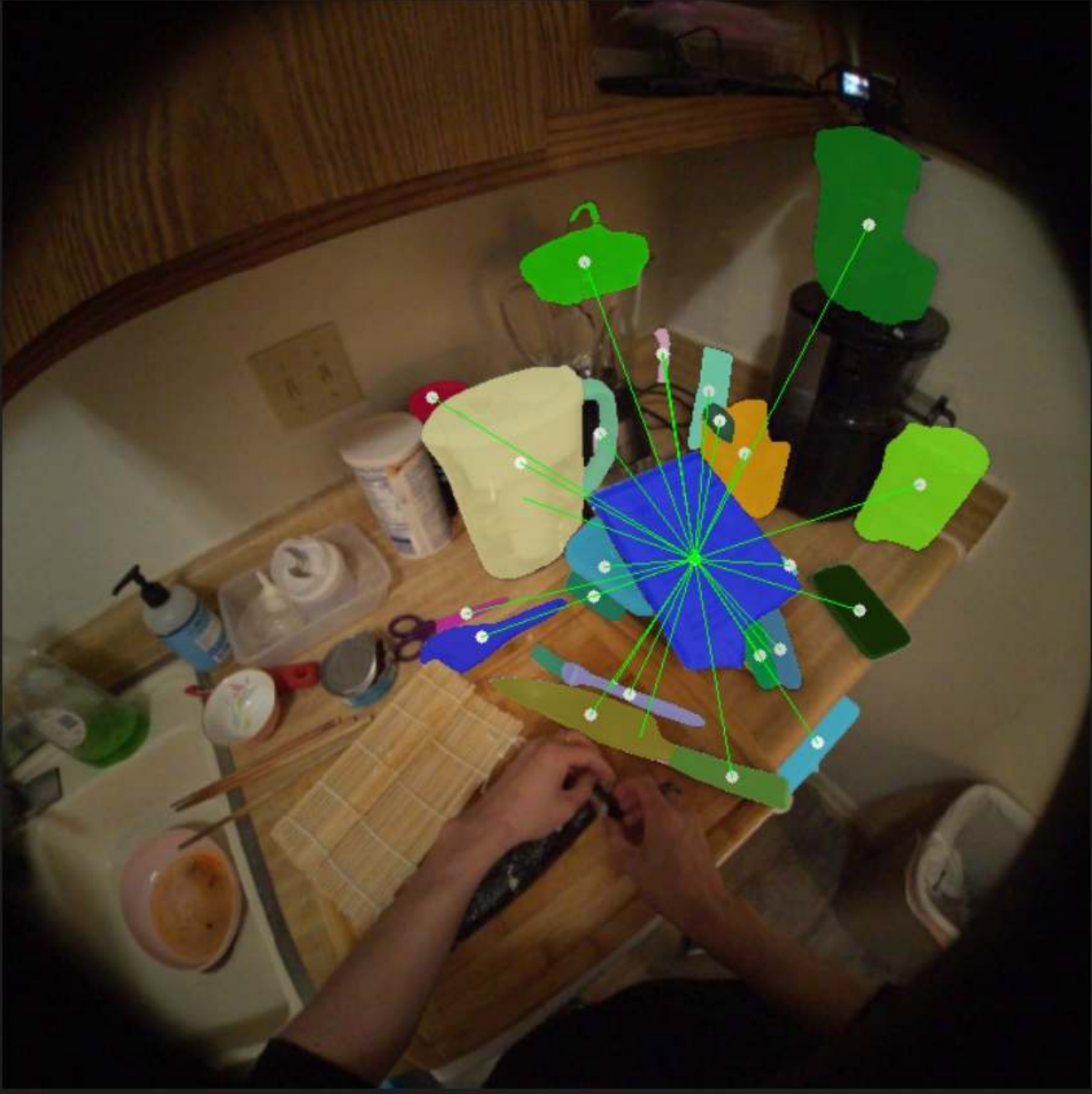}
    \includegraphics[width=0.64\linewidth]{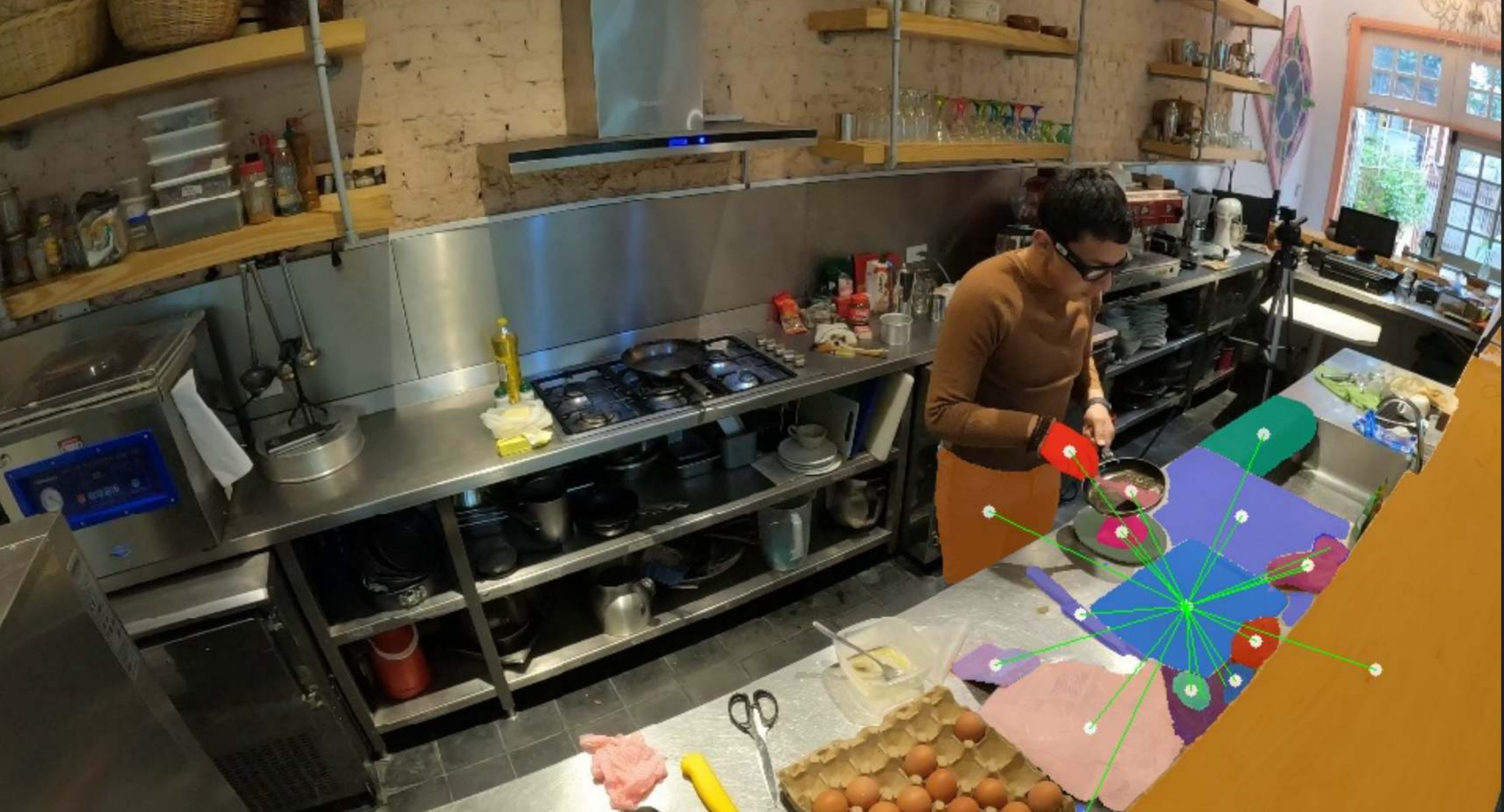}
    
    \caption{\textbf{Hard Negatives mining examples.} We visualize $2^{nd}$ order adjacent neighbors both in ego (left) and exo (right) scenarios.}
    \label{fig:hard negs}
\end{figure}

While the object embedding $\mathbf{o}_{n}$ contains very discriminative object features, the context embedding $\mathbf{c}_{n}$ incorporates surrounding information to help localizing the object in the other view, but this surrounding context also introduces ambiguity in cluttered environments, where nearby objects share a similar context. 
To address this, we introduce a hard-negative mining strategy based on adjacent neighbors, encouraging the model to disambiguate between nearby but distinct objects with similar context.
In the destination view, we construct a graph of mask segments based on the pixel centers of each mask using the Delaunay Triangulation, as \cref{fig:hard negs} shows. This results in a binary adjacency matrix $\mathcal{A} \in \{0,1\}^{N \times N}$, defining the connectivity between segments. We define $\mathcal{N}({\mathbf{o}_n})$ to the set of neighbors of object $\mathbf{o}_{n}$. Then, we take the second order neighbor set $\mathcal{N}^2({\mathbf{o}_n}) = \{\mathcal{N}({\mathbf{o}_j}) \;\; \forall \mathbf{o}_j \in \mathcal{N}({\mathbf{o}_n})\}$. Finally, we consider the joint set of first and second order neighbors as the hard negative candidates $\mathcal{O}^{-}_{n} = \{\mathcal{N}({\mathbf{o}_n}) \cup \mathcal{N}^2({\mathbf{o}_n})\}$.

\subsection{Ego$\leftrightarrow$Exo Cross Attention}
\label{sec:cross_attn}

Although the mask context embedding incorporates surrounding contextual information, it lacks a global representation across views. 
Therefore, we introduce a \textit{Ego$\leftrightarrow$Exo Cross Attention} mechanism, which enhances the object embedding by extracting its corresponding semantic features in the other view.
Specifically, we compute a cross attention operation \cite{vaswani2017attention} between the candidate object masks $\mathbf{o}_{n}$ and the source image feature map $\psi(I^{S})$.
\begin{equation*}
\hat{\mathbf{o}}_{n} = \text{Softmax} \left( \frac{ \mathbf{o}_{n} W_Q  \cdot  ( \psi\left(I^{S}\right) W_K)^\top}{\sqrt{d}} \right) \cdot \psi\left(I^{S}\right)W_V
\end{equation*}
We compute query vectors from the candidate object descriptors $\mathbf{o}_{n}$ with a linear projection $W_Q$, while key and value vectors represent the overall source image features $\psi(I^{S})$ using the $W_K$ and $W_V$ linear layers. Before the cross-attention operation, we incorporate a learnable positional embedding to encode the spatial location of the patch tokens and a standard Layer Norm operation. Intuitively, the cross-view embedding of the object-mask candidates $\hat{\mathbf{o}}_{n}$ captures how each potential mask candidate is represented in the source view.
Similarly, we compute the cross-view embedding of the source mask $\hat{\mathbf{o}}_{S}$ using the source mask descriptor $\mathbf{o}_{S}$ for the queries and the overall destination image features $\psi(I^D)$ for the keys and values.

\subsection{Mask Matching Contrastive Loss}
\label{sec:loss}

The final descriptor $\rho_{n}$ is obtained from the $n$-th candidate mask by concatenating its refined cross-view embedding $\hat{\textbf{o}}_{n}$, the context embedding $\textbf{c}_{n}$ and the object embedding $\textbf{o}_{n}$; while the final descriptor $\rho_{S}$ of the source mask is the result of concatenating $\hat{\textbf{o}}_{S}$, $\textbf{c}_{S}$ and $\textbf{o}_{S}$, respectively.
Then, a shallow multi-layer perceptron $f_\theta(x) \in \mathbb{R}^{d_f}$ maps the cross-view embeddings to a latent feature representation, 
where $d_f$ is the dimension of the common feature space. 
This mapping ensures that both egocentric and exocentric masks are embedded within a shared latent space, enabling a cross-view comparison.

Our contrastive loss is based on InfoNCE \cite{oord2018representation}. We select a batch $\mathcal{B}$ of $|\mathcal{B}|$ elements, one positive and $|\mathcal{B}|-1$ negatives from the list of closest neighbors around the target object in the other view $\mathcal{O}^{-}_{n}$ as defined in Section \ref{sec:mining}. If the number of neighbors is greater than the negative batch size, $|\mathcal{O}^{-}_{n}| > |\mathcal{B}|$ we randomly select a subset of $|\mathcal{B}|$ elements from the neighbor set $\mathcal{O}^{-}_{n}$. If $|\mathcal{O}^{-}_{n}| < |\mathcal{B}|$ we also include masks from random objects from the rest of the image. If the number of segmented objects is less than the neighbor set $\mathcal{O}^{-}_{n}$, the objects are duplicated in order to fill the negative batch size $|\mathcal{B}|$.
Finally, we apply the pairwise cosine similarity $\text{sim}(\cdot, \cdot)$ between the source mask embedding $f_\theta(\rho_S)$ and the batch of $|\mathcal{B}|$ mask candidates embeddings $\{ f_\theta(\rho_n) \}_{n=1}^{|\mathcal{B}|}$ for computing the training loss:
\begin{equation*}
    \mathcal{L}_{M}(\rho^+, \rho_{S}) = - \log \frac{\exp(\text{sim}(f_\theta (\rho^+), f_\theta (\rho_{S}))/\tau)}{\sum^{|\mathcal{B}|}_{n=1} \exp(\text{sim}(f_\theta(\rho_{n}), f_\theta(\rho_{S}))/\tau)}
\end{equation*}
where $f_\theta (\rho^+)$ is the positive element in the batch. This \emph{mask matching contrastive loss} aligns the corresponding cross-view object embeddings while it separates the remaining object candidates in the shared feature space.

\subsection{Inference}

At inference, we choose the object candidate whose embedding is closest to the source object in the latent space.
\begin{equation*}
    \mathcal{M}_{n^*}, \; \text{where} \;\; n^* = \arg\max (\text{sim} \left[ f_\theta(\rho_{n}), f_\theta(\rho_{S})\right])
\end{equation*}
Intuitively, the similarity score should be low for objects that do not share context or appearance.

\section{Experiments}
\label{sec:results}

\subsection{Experimental Setup}

\begin{table*}
    \centering
    \resizebox{0.90\textwidth}{!}{
    \begin{tabular}{l ccc ccc c cc} 
         & \multicolumn{3}{c}{Ego2Exo}&  \multicolumn{3}{c}{Exo2Ego} & & \multicolumn{2}{c}{Num. Param. (M)} \\ 
         \cmidrule(lr){2-4} \cmidrule(lr){5-7} \cmidrule(lr){9-10}
         Method & IoU$\uparrow$ &  Loc.E$\downarrow$ &  Cont.A$\uparrow$ & IoU$\uparrow$ & Loc.E$\downarrow$ &  Cont.A$\uparrow$ & Total-IoU$\uparrow$  & Total & Train\\ \hline  
         PSALM \cite{zhang2024psalm} (Zero-shot) &  7.4 & 0.266 & 0.121 & 2.1 & 0.294  & 0.058 & 4.8 & 1587.1& 0\\  
         CMX \cite{zhang2023cmx} & 6.8 & 0.110 & 0.137 & 12.0 & 0.166 & 0.177 & 9.4 & 138.0 & 17.3 \\ 
         XSegTx \cite{grauman2024ego} &  18.9&  0.070&  0.386&  27.1& 0.104 & 0.358 & 23.0 & 12.1& 3.6\\   
         XMem \cite{grauman2024ego} &  19.3&  0.151&  0.262&   16.6 & 0.160& 0.240 & 18.0 & 62.2& 62.2\\  
         XMem + XSegTx \cite{grauman2024ego} &  34.9 & 0.038 & 0.559& 25.0 &  0.117& 0.237 & 30.0 & 75.6& 67.1\\ \hline
         Ours (k-NN baseline)& 31.9 & 0.195& 0.414 & 30.9 & 0.127 & 0.373 & 31.4 & 154.0 &0 \\ 
         Ours (O-MaMa)& \textbf{42.6} &\bf{0.033} &\textbf{0.590}& \textbf{44.1} & \textbf{0.082} & \textbf{0.524} & \textbf{43.4} & 165.6 & 11.6 \\  \hline

    \end{tabular}
    }
    \caption{\textbf{Results on the Ego-Exo4D Correspondences v2 test split.}}
    \label{tab:state_of_art}
\end{table*}

\begin{table}[ht]
    \centering
    \resizebox{\columnwidth}{!}{%
    \begin{tabular}{l cc c cc}
        Method & \makecell{Ego2Exo\\IoU$\uparrow$} & \makecell{Exo2Ego\\IoU$\uparrow$}  & \makecell{Total\\IoU$\uparrow$} &
        \makecell{Total\\Param.(M)} & \makecell{Train\\Param.(M)}\\ \hline
        XSegTx & 6.2 & 30.2 & 18.2 & 12.1 & 3.6 \\
        XMem & 17.2 & 20.7 & 19.0 & 62.2 & 62.2 \\
        XMem + XSegTx & 36.9 & 36.1 & 36.5 & 75.6 & 67.1 \\
        PSALM (zero-shot) & 7.9 & 9.6 & 8.8 & 1587.1 & 0 \\
        PSALM (fine-tuned) & 41.3 & 44.1 & 42.7 & 1587.1 & 1587.1 \\
        ObjectRelator & 44.3 & 50.9 & 47.6  & 1587.3 & 1587.3 \\ \hline
        Ours (k-NN baseline) & 40.5 & 40.6 & 40.6 & 154.0 & 0 \\
        Ours (O-MaMa) & \textbf{50.1} & \textbf{54.2} & \textbf{52.1} & 165.6 & 11.6 \\ \hline
    \end{tabular}
    }
    \caption{\textbf{Ego-Exo4D Correspondences v1 val split results.}}
    \label{tab:sota_val_v1}
\end{table}

\begin{table*}
    \centering
    \resizebox{\textwidth}{!}{
    \begin{tabular}{c ccccc ccc ccc c}
                &  & & &  & & \multicolumn{3}{c}{Ego2Exo}&  \multicolumn{3}{c}{Exo2Ego}\\ 
               \cmidrule(lr){7-9} \cmidrule(lr){10-12}
               Exp. &$\mathcal{L}_{M}$&Context&Adj. Neg &C.Attn &Global Union&  IoU$\uparrow$& 
               Loc.E$\downarrow$&  Cont.A $\uparrow$ & IoU$\uparrow$ & 
               Loc.E$\downarrow$ & Cont.A$\uparrow$ &
               Total-IoU$\uparrow$\\ \hline
               Base. &-&-& -& -&-&  35.2&  
               0.191& 0.455& 34.9&
               0.163&0.423 & 35.1\\ \hline
               A &\checkmark& -& -& -& -& 42.2 & 
               0.074 & 0.571 & 44.7& 
               0.112&0.546& 43.5\\
               B &\checkmark&\checkmark& -& -& Concat &  42.7& 
               0.069&   0.577& 44.4& 
               0.116&0.543&43.6\\
               C &\checkmark&\checkmark& \checkmark&  -& Concat & 46.9& 
               0.079&0.599& 45.6&
               0.107&0.548&46.3\\
              D &\checkmark&\checkmark& \checkmark& \checkmark & Weighted Sum.& 47.3& 
              0.064& 0.611& 46.8&
              0.112&0.543&47.1\\ 
              E &\checkmark&\checkmark& \checkmark& \checkmark & Concat & \textbf{48.3}& 
              \textbf{0.062}& \textbf{0.621}& \textbf{49.6}&
              \textbf{0.101}&\textbf{0.576} & \textbf{49.0}\\ \hline
      &\multicolumn{5}{c}{Relative Gain $\%$ of x with respect to y $\frac{(x-y)}{y}$}& \textcolor{ForestGreen}{+37.2$\%$}& 
      \textcolor{ForestGreen}{+67.5$\%$} & \textcolor{ForestGreen}{+36.5$\%$} & \textcolor{ForestGreen}{+42.1$\%$} & 
      \textcolor{ForestGreen}{+38.0$\%$} & \textcolor{ForestGreen}{+36.2$\%$}& \textcolor{ForestGreen}{+39.6$\%$}\\ \hline
    \end{tabular}
    }
    \caption{\textbf{Ablation study on the O-MaMa proposed modules on the 10$\%$ of the validation set.}}
    \label{tab:arch ablate}
\end{table*}

\noindent
\textbf{Training dataset.} 
We use the novel Ego-Exo4D \cite{grauman2024ego} dataset for our experiments. Ego-Exo4D is a massive-scale multi-modal video dataset containing synchronized egocentric and exocentric recordings of human activities.
Specifically, we consider the Ego-Exo4D Correspondences set, which includes 1.8M synchronized object masks annotated at 1 FPS, covering 5.6K objects across 1335 unique videos and six different activities (\textit{cooking, bike repair, health, music, basketball,} and \textit{soccer}).

\noindent
\textbf{Evaluation.}
Following the official Ego-Exo4D Correspondences benchmark \cite{grauman2024ego}, we adopt the Intersection over Union (IoU) as the primary evaluation metric.
We also report
the Contour Accuracy (Cont.A) \cite{perazzi2016benchmark} to measure the similarity between predicted and ground-truth mask contours after translation, and the Location Error (Loc.E), which quantifies the normalized distance between the predicted and ground-truth mask centroids. 

\noindent
\textbf{Implementation details.}
We employ FastSAM \cite{zhao2023fast} with default hyper-parameters (0.9 IoU, 0.4 confidence score) to generate dense candidate masks in the destination view. 
FastSAM achieves performance comparable to SAM \cite{kirillov2023segment} while being 50$\times$ faster and utilizing just 68M parameters.
For feature extraction, we use a DINOv2 \cite{oquab2023dinov2} ViT-B/14 model, which consists of 86M parameters. 
Our model is trained with the AdamW optimizer \cite{loshchilov2017decoupled} and an initial learning rate of 8$\cdot 10^{-5}$ with cosine annealing scheduling. We use a batch size of 24 image pairs, sampling 32 masks candidates in each destination image during training.  
We conduct our experiments on two NVIDIA GeForce RTX 4090.

\begin{figure}
    \centering
    \includegraphics[width=0.99\linewidth]{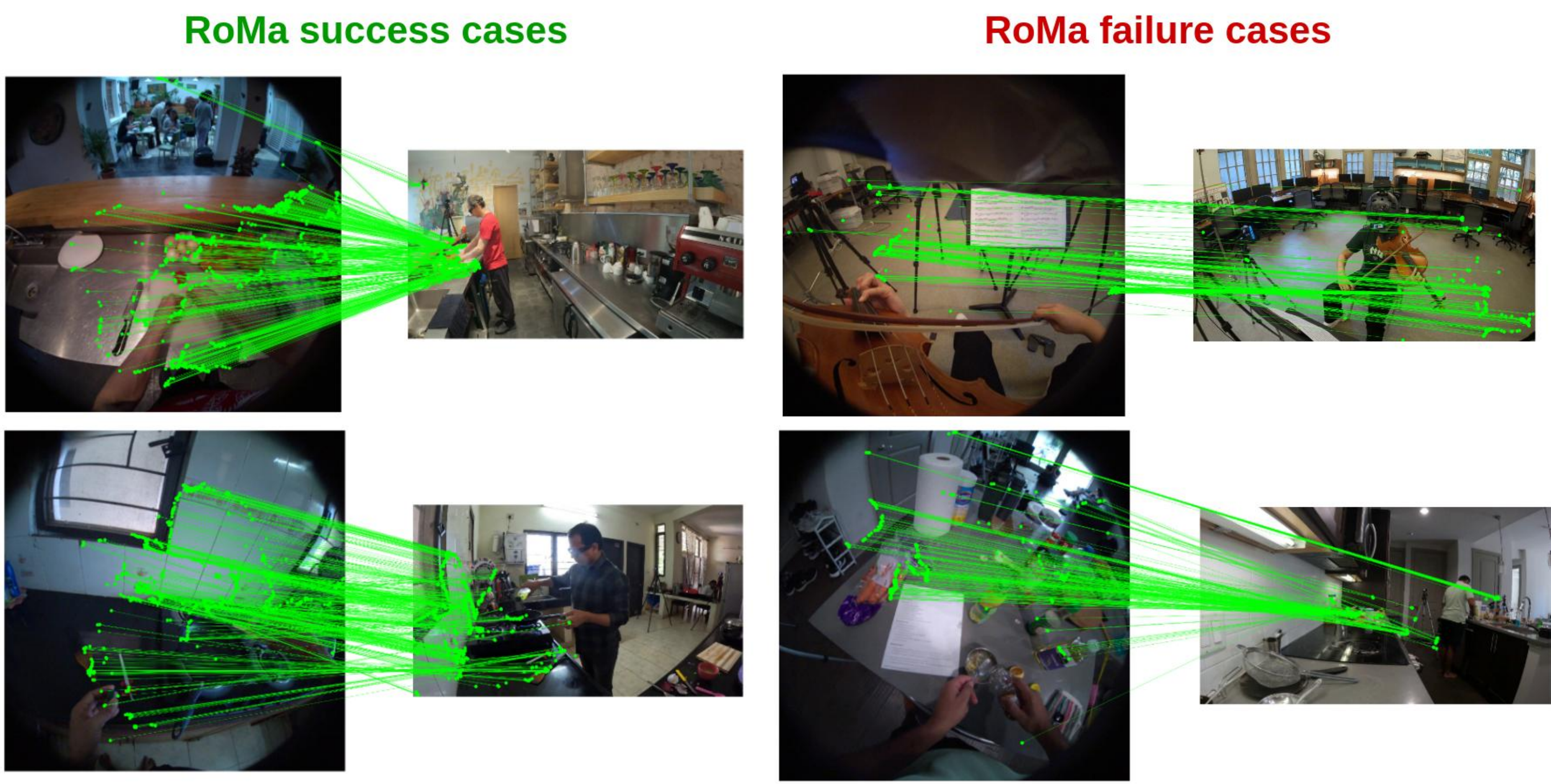}
    \caption{\textbf{RoMa \cite{edstedt2024roma} success and failure cases.} The extreme view variance makes that, even SOTA methods in geometry matching like RoMa \cite{edstedt2024roma}, fail in extracting matches.}
    \label{fig:geometry}
\end{figure}

\begin{table*}[ht]
\centering
\label{tab:results_comparison}
\resizebox{0.90\textwidth}{!}{
\begin{tabular}{l c c ccc ccc c}
 & & & \multicolumn{3}{c}{Ego2Exo} & \multicolumn{3}{c}{Exo2Ego} \\
\cmidrule(lr){4-6} \cmidrule(lr){7-9}
Method & Geometry & $\mathcal{L}_{M}$ & IoU$\uparrow$ & Loc.E$\downarrow$ & Cont.A$\uparrow$ & IoU$\uparrow$ & Loc.E$\downarrow$ & Cont.A$\uparrow$ & Total-IoU$\uparrow$\\
\hline
\multirow{2}{*}{Max-Pool($\mathbf{b}$)-DINOv2 \cite{oquab2023dinov2}} & \xmark & \xmark & 5.8 & 0.306 & 0.119 & 17.8 & 0.216 & 0.253 & 11.8 \\
& \cmark & \xmark & 6.9 & 0.302 & 0.132 & 20.2 & 0.210 & 0.278 & 13.6 \\
\hline
\multirow{2}{*}{Centroid($\mathcal{M}$)-DINOv2 \cite{oquab2023dinov2}} & \xmark & \xmark & 25.6 & 0.202 & 0.357 & 24.1 & 0.178 & 0.326 & 24.9 \\
& \cmark & \xmark & 26.5 & 0.190 & 0.378 & 26.0 & 0.172 & 0.346 & 26.3\\
\hline
\multirow{3}{*}{Avg-Pool($\mathbf{b}$)-DINOv2 \cite{oquab2023dinov2}} & \xmark & \xmark & 21.8 & 0.245 & 0.324 & 21.2  & 0.201 & 0.291 & 21.5 \\
& \cmark & \xmark & 23.2 & 0.238 & 0.345 & 23.7 & 0.195 & 0.314 & 23.5\\
& \xmark & \cmark & 27.8 & 0.092 & 0.426 & 44.1 & \textbf{0.111} & 0.537 & 36.0 \\
\hline
\multirow{3}{*}{Avg-Pool($\mathbf{b}$)-CLIP \cite{radford2021learning}} & \xmark & \xmark & 24.5 & 0.257 & 0.325 & 23.9 & 0.234 & 0.301 & 24.2 \\
& \cmark & \xmark & 26.2 & 0.220 & 0.359 & 26.7 & 0.209 & 0.335 & 26.5\\
& \xmark & \cmark & 27.5 & 0.170 & 0.379 & 40.4 & 0.155 & 0.477 & 34.0\\
\hline
\multirow{3}{*}{Avg-Pool($\mathcal{M}$)-DINOv2 \cite{oquab2023dinov2}} & \xmark & \xmark & 35.2 & 0.191 & 0.455 & 34.9 & 0.163 & 0.423 & 35.1\\
& \cmark & \xmark & 35.4 & 0.184 & 0.467 & 36.6 & 0.156 & 0.440 & 36.0\\
& \xmark & \cmark & \textbf{42.2} & \textbf{0.074} & \textbf{0.571} & \textbf{44.7} & 0.112 & \textbf{0.546} & \textbf{43.5}\\
\hline
\end{tabular}
}
\caption{\textbf{Ablation study on the mask descriptors and the influence of learning and geometry constraints.} We compare the effects of leveraging inferred camera pose constraints or training a simple MLP with our $\mathcal{L}_{M}$, configuration that corresponds to Exp.A in \cref{tab:arch ablate}.}
\label{tab:no_learning_ablation}
\end{table*}

\subsection{Baseline Models}

We compare our approach against the following baselines\footnote{See the supplementary material for implementation details}:

\begin{itemize}
    \item \textbf{XSegTx and XView-XMem} are the official baselines \cite{grauman2024ego}. XSegTx adapts an image co-segmentation model \cite{shen2022learning}, extended with a cross-view temporal memory \cite{cheng2022xmem}. 
    \item \textbf{CMX} \cite{zhang2023cmx} is a transformer-based segmentation model that fuses two modalities. 
    We concatenate the query mask with the source image, and we adapt the decoder to predict the mask.
    \item \textbf{PSALM} \cite{zhang2024psalm} combines a LLM with Mask2Former \cite{cheng2022masked} to perform zero-shot segmentation. \textbf{ObjectRelator} \cite{fu2024objectrelator} trains PSALM with specialized cross-view modules. 
    \item \textbf{k-Nearest Neighbors (k-NN)}. This is a naïve version of our approach. 
    We extract descriptors of the generated mask candidates in the destination view, and we select the most similar to the query mask in the source view. 
    \item \textbf{Geometry Methods} \cite{edstedt2024roma}. We restrict the k-NN search to only the masks that satisfy the epipolar line restriction, in order to evaluate the geometrical constraints of traditional geometrical matching methods. We tried LightGlue \cite{lindenberger2023lightglue} using either SuperGlue \cite{sarlin2020superglue} (43.8 $\%$ success rate\footnote{Rate of success cases (see~\cref{fig:geometry}) when extracting matches in pairs of images. Refer to the supplementary material for more details.}), SIFT \cite{lowe2004distinctive} (42 $\%$ success rate), DISK \cite{tyszkiewicz2020disk} (31.2 $\%$ success rate) or ALIKED \cite{zhao2023aliked} (40.2 $\%$ success rate),  and compared them using RoMa \cite{edstedt2024roma} (67.6 $\%$ success rate). Due to its more view variance robustness, we select this last method to obtain the fundamental matrix and transfer the mask centroid in the source view to its epipolar line in the destination view, and we discard those candidate masks further than a certain threshold to the epipolar line.
\end{itemize}

\subsection{Comparison with the State of the Art}

\cref{tab:state_of_art} presents results on the EgoExo4D Correspondences v2 test set, demonstrating the our approach's effectiveness. 
Even our simplest version, the k-NN baseline, already surpasses the official XMem+XSegTx, achieving 31.9 IoU in Ego2Exo and 30.9 IoU in Exo2Ego tasks.
Our full method, O-MaMa, further improves performance, reaching 42.6 Ego2Exo and 44.1 Exo2Ego IoU, representing considerable relative gains\footnote{We compute the relative gain\% of $x$ relative to $y$ as $100 \cdot (\frac{x-y}{y})$.} of up to +22.1 $\%$ and +76.4 $\%$ over XMem+XSegTx.
The improvement is consistent in the other metrics, where O-MaMa obtains 0.033 Loc.E, 0.590 Cont.A in the Ego2Exo task and 0.082 Loc.E and 0.524 Cont.A in the Exo2Ego task.

As ObjectRelator \cite{fu2024objectrelator} reports only results on the outdated EgoExo4D Correspondences v1 validation split, we also show results in this split in \cref{tab:sota_val_v1}.
O-MaMa also achieves the best performance, obtaining a 50.1 Ego2Exo IoU and 54.2 Exo2Ego IoU, while requiring only 1 $\%$ of the trainable parameters compared to \cite{fu2024objectrelator}.
The potential of our approach is further demonstrated by comparing our k-NN baseline with PSALM \cite{zhang2024psalm} in zero-shot inference.
While PSALM achieves only 7.9 and 9.6 IoU in the Ego2Exo and Exo2Ego tasks, respectively, our k-NN baseline scores 40.5 and 40.6 IoU while using approximately 10$\%$ of the total number of parameters.
This highlights the underlying challenges of the cross-view segmentation task and showcases that reformulating the problem as an object mask matching task significantly boosts zero-shot performance.

\subsection{Ablation study}

\begin{figure}
    \centering
    \includegraphics[width=0.99\linewidth]{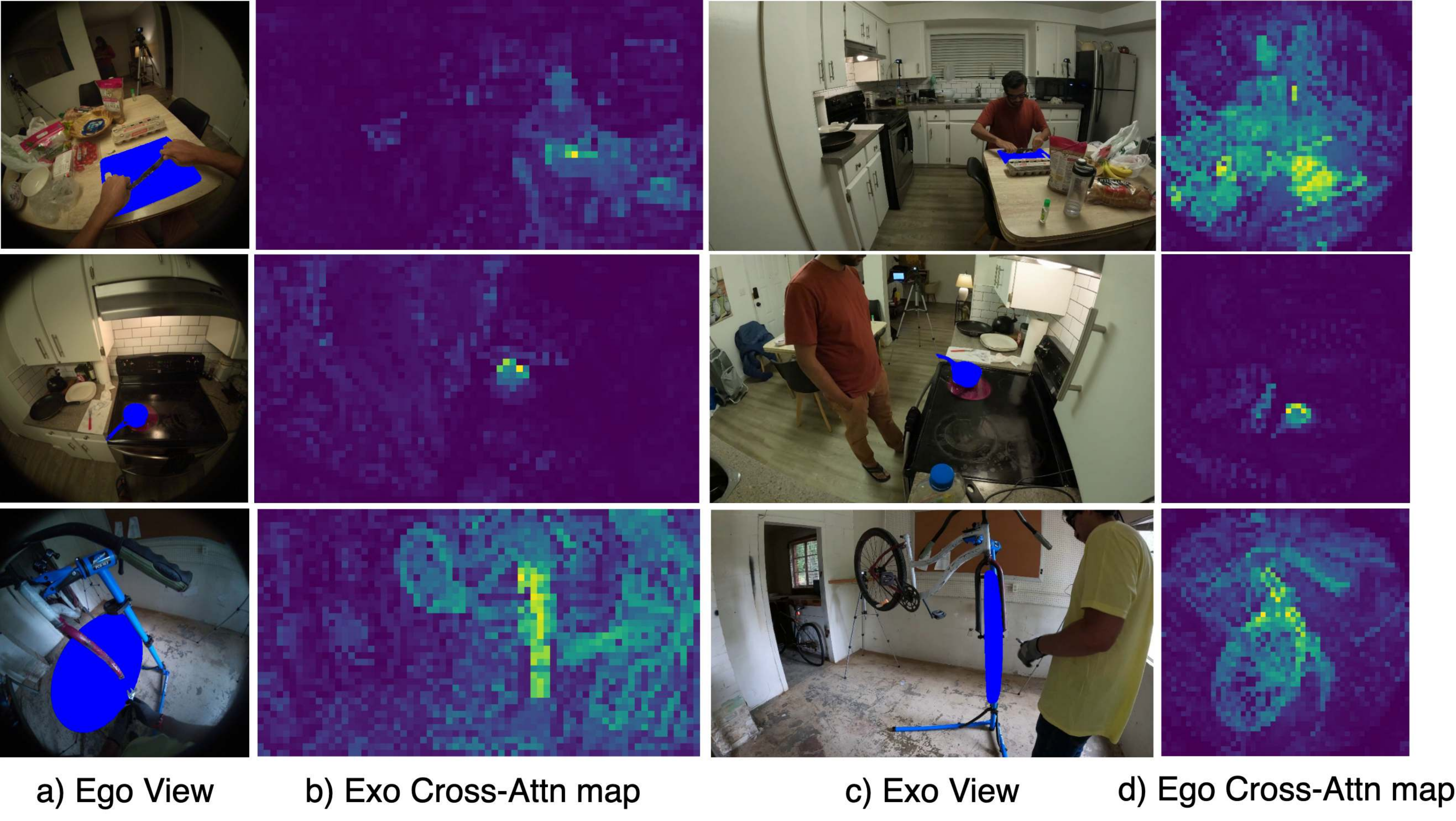}
    \caption{\textbf{Ego$\leftrightarrow$Exo Cross-Attention maps}}
    \label{fig:attn maps}
\end{figure}

\noindent
\textbf{O-MaMa architecture.}
\cref{tab:arch ablate} details the contribution of each O-MaMa component.
Experiment A highlights the benefits of training a simple MLP with our novel Mask Matching Contrastive Loss $\mathcal{L}_{M}$, which aligns cross-view embeddings in a common latent space and improves IoU from 35.2 to 42.2 (Ego2Exo) and 34.9 to 44.7 (Exo2Ego). 
Second, Experiments A, B and C show that incorporating regional context is only beneficial when we sample adjacent negatives during training, as the hard-mining strategy forces the model to learn more fine-grained discriminative embeddings in nearby candidates with similar context but different mask descriptor.
Next, our Ego$\leftrightarrow$Exo Cross Attention mechanism introduces cross-image content and global information into the object embedding.
As \cref{fig:attn maps} shows, this module incorporates the object features from the other perspective, smoothing the cross-view alignment and improving the final performance.
Finally, the joined effect of all our proposed modules specially improves the Loc.E, with relative improvements of +67.5 $\%$ (Ego2Exo) and +38.0 $\%$ (Exo2Ego), which yields a final gain of +37.2 $\%$ Ego2Exo and +42.1 $\%$ Exo2Ego IoU.
This demonstrates that, while the k-NN baseline is agnostic to the candidate mask location,
(it just selects the most similar match),

\begin{figure}
    \centering
    \includegraphics[width=0.99\linewidth]{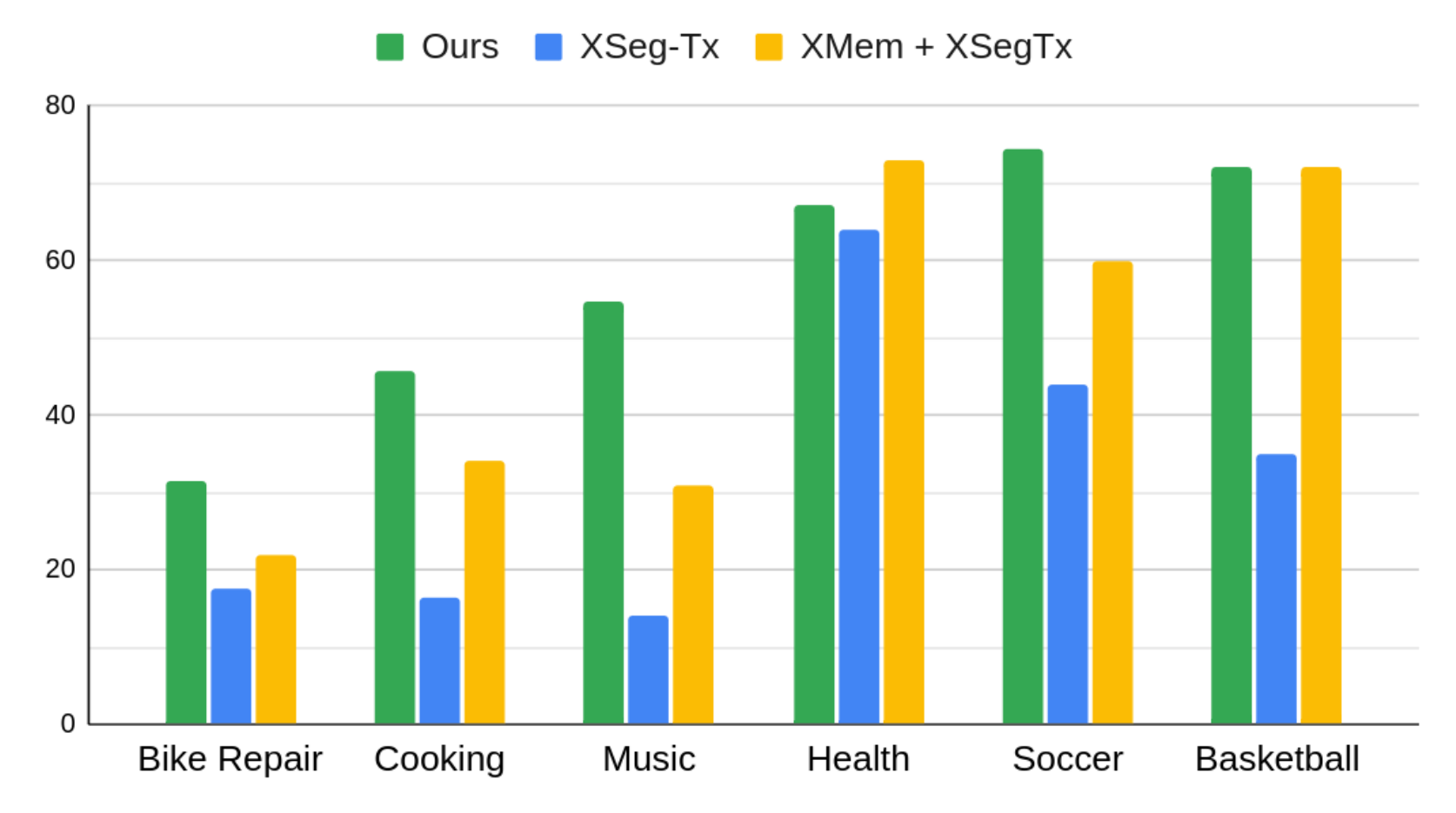}
    \caption{\textbf{Per-tasks Ego2Exo IoU performance.}}
    \label{fig:iou ego exo}
\end{figure}

\begin{figure*}
    \centering
    \includegraphics[width=0.95\linewidth]{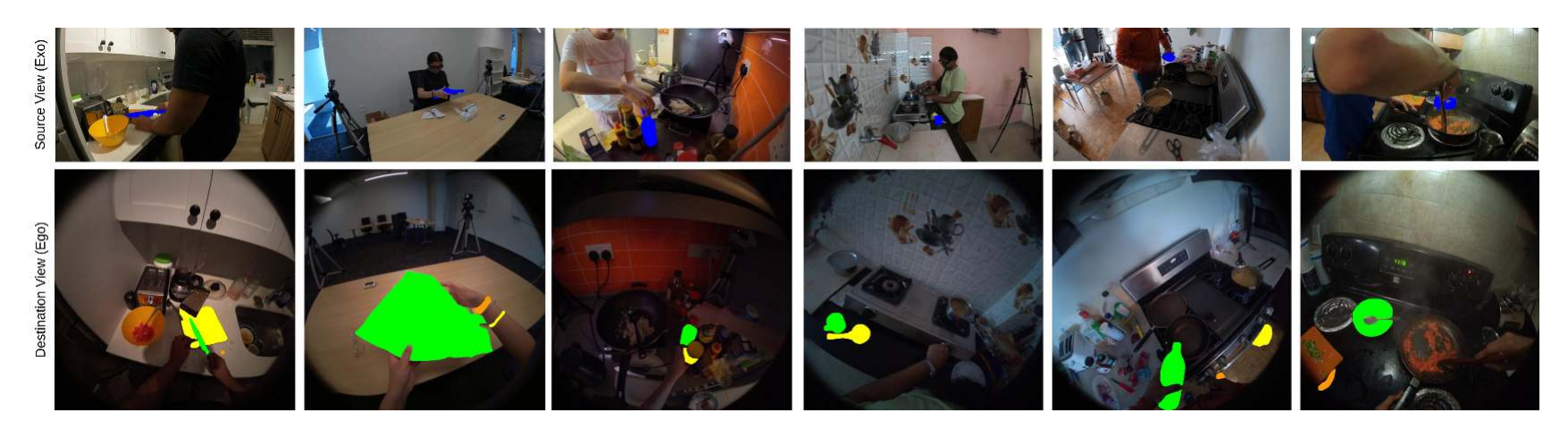}
    \caption{\textbf{Exo2Ego Qualitative Results.} We show the source mask in \textcolor{blue}{blue} and the top 3 target masks in \textcolor{green}{green}, \textcolor{yellow}{yellow} and \textcolor{orange}{orange}.}
    \label{fig:qualitative2}
\end{figure*}

\begin{figure}
    \centering
    \includegraphics[width=0.99\linewidth]{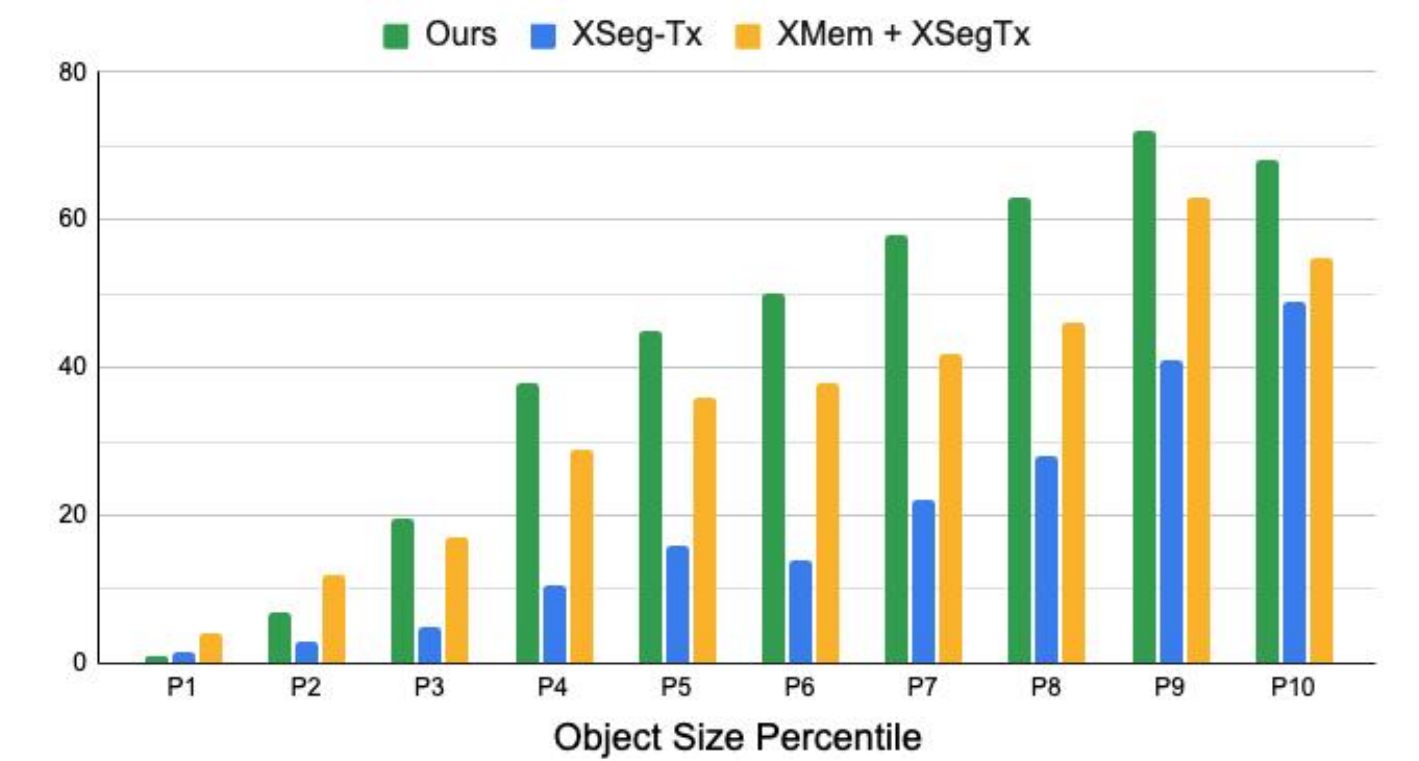}
    \caption{\textbf{Ego2Exo IoU performance across different object sizes in the destination view.}}
    \label{fig:IOU_PERCENTILE}
\end{figure}

\noindent
\textbf{Mask Descriptors.}
\cref{tab:no_learning_ablation} compares different pooling strategies for obtaining a mask descriptor. 
The k-NN baseline, which relies solely on object semantic similarity, shows that averaging DINOv2 upsampled features over mask pixels provides the best results (35.2 and 34.9 IoU), as it retains the fine-grained object representation from the dense DINOv2 feature map. 
This strategy outperforms CLIP-based descriptors (24.5 Ego2Exo and 23.9 Exo2Ego IoU), DINOv2 average pooling over the bounding box (21.8 Ego2Exo and 21.2 Exo2Ego IoU) or DINOv2 mask centroid (25.6 Ego2Exo and 24.1 Exo2Ego IoU).
\cref{tab:no_learning_ablation} also reports that applying geometric constraints yields a minor performance gain (35.2 vs. 35.4 Ego2Exo IoU, and 34.9 vs. 36.6 Exo2Ego IoU when pooling DINOv2 mask features) due to the low success rate of camera pose estimation methods. 
As \cref{fig:geometry} shows, even RoMa \cite{edstedt2024roma} struggles with the high viewpoint variance between ego and exo perspectives.
This improvement is even less significant when compared to learning view-invariant features with $\mathcal{L}_{M}$ (35.2 vs. 42.2 Ego2Exo IoU and 34.9 vs. 44.7 Exo2Ego IoU), highlighting the need to extract stronger visual cues.

\noindent
\textbf{Detailed performance per task and mask size.}
\cref{fig:iou ego exo} shows the IoU across different scenarios in the Ego2Exo task, where O-MaMa outperforms XMem + XSeg-Tx in most cases, including the challenging \textit{cooking} and \textit{bike repair} activities, which involve cluttered environments and objects of multiple sizes.
\cref{fig:IOU_PERCENTILE} analyzes the segmentation performance across the target mask sizes, showing that O-MaMa excels in medium and large-size objects.
However, it still struggles with very small objects, as extracting a meaningful mask descriptor remains challenging.

\subsection{Qualitative results}

\begin{figure}
    \centering
    \includegraphics[width=0.99\linewidth]{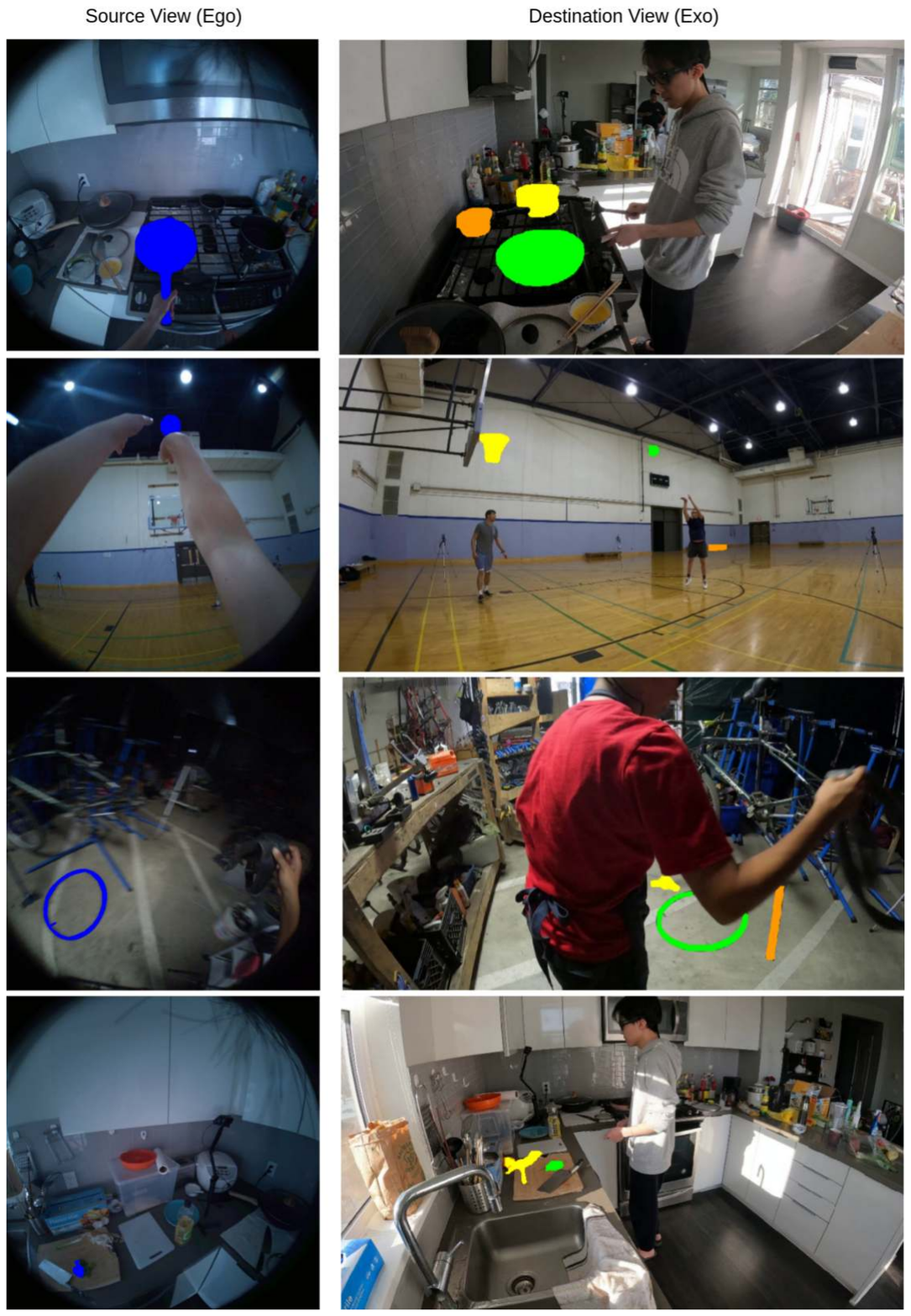}
    \caption{\textbf{Ego2Exo Qualitative Results.} For visualization purposes, we show the top 3 masks in \textcolor{green}{green}, \textcolor{yellow}{yellow} and \textcolor{orange}{orange}.}
    \label{fig:qualitative1}
\end{figure}

We show qualitative examples for the Exo2Ego (\cref{fig:qualitative2}) and Ego2Exo (\cref{fig:qualitative1}) tasks.
The results show that top mask candidates are closely aligned due to their similar context, but our method correctly matches the top-1 mask candidate with the target object.
FastSAM’s fine-grained zero-shot capabilities yield high-quality segmentation masks (\eg, the \textit{tire} in \cref{fig:qualitative1} and the \textit{knife} or \textit{bottle} in \cref{fig:qualitative2}). 
However, as a limitation of our approach, they may produce partial segmentations when they capture only a part of the object (\eg, the \textit{saucepan} in \cref{fig:qualitative1}).
Finally, the total inference time of our approach is 250ms on average, of which 70ms correspond to the FastSAM mask extraction.

\section{Conclusions}
\label{sec:conclusion}

In this work, we address the problem of ego-exo object correspondences, a key step for multi-agent perception. 
We demonstrate that reformulating cross-view segmentation as an object mask matching problem simplifies the task while improving accuracy under zero-shot conditions.
Our Mask Matching Contrastive Loss effectively aligns cross-view embeddings, while DINOv2 pooled mask features preserve fine-grained details.
The proposed Hard Negative Adjacent Mining strategy enhances object differentiation, and Ego$\leftrightarrow$Exo Cross Attention integrates global cross-view context.
As a result, O-MaMa achieves state-of-the-art performance on the EgoExo4D Correspondences task while using considerably fewer parameters, obtaining a unified fine-grained segmentation and strong cross-view understanding.

\section{Acknowledgments}
This work was supported by projects PID2021-125209OB-I00 and TED2021-129410B-I00, (MCIN/AEI/10.13039/501100011033 and FEDER/UE and NextGenerationEU/PRTR),  and Aragon Government DGA T45-23R.

{
    \small
    \bibliographystyle{ieeenat_fullname}
    \bibliography{main}

\begin{thebibliography}{90}
\providecommand{\natexlab}[1]{#1}
\providecommand{\url}[1]{\texttt{#1}}
\expandafter\ifx\csname urlstyle\endcsname\relax
  \providecommand{\doi}[1]{doi: #1}\else
  \providecommand{\doi}{doi: \begingroup \urlstyle{rm}\Url}\fi

\bibitem[Bansal et~al.(2022)Bansal, Arora, and Jawahar]{bansal2022my}
Siddhant Bansal, Chetan Arora, and CV Jawahar.
\newblock My view is the best view: Procedure learning from egocentric videos.
\newblock In \emph{European Conference on Computer Vision}, pages 657--675. Springer, 2022.

\bibitem[Bar(2004)]{bar2004visual}
Moshe Bar.
\newblock Visual objects in context.
\newblock \emph{Nature Reviews Neuroscience}, 5\penalty0 (8):\penalty0 617--629, 2004.

\bibitem[B{\"a}rmann and Waibel(2022)]{barmann2022did}
Leonard B{\"a}rmann and Alex Waibel.
\newblock Where did i leave my keys?-episodic-memory-based question answering on egocentric videos.
\newblock In \emph{Proceedings of the IEEE/CVF Conference on Computer Vision and Pattern Recognition}, pages 1560--1568, 2022.

\bibitem[Bay et~al.(2008)Bay, Ess, Tuytelaars, and Van~Gool]{bay2008speeded}
Herbert Bay, Andreas Ess, Tinne Tuytelaars, and Luc Van~Gool.
\newblock Speeded-up robust features (surf).
\newblock \emph{Computer vision and image understanding}, 110\penalty0 (3):\penalty0 346--359, 2008.

\bibitem[Bolya et~al.(2019)Bolya, Zhou, Xiao, and Lee]{bolya2019yolact}
Daniel Bolya, Chong Zhou, Fanyi Xiao, and Yong~Jae Lee.
\newblock Yolact: Real-time instance segmentation.
\newblock In \emph{Proceedings of the IEEE/CVF international conference on computer vision}, pages 9157--9166, 2019.

\bibitem[Caelles et~al.(2019)Caelles, Pont-Tuset, Perazzi, Montes, Maninis, and Van~Gool]{caelles20192019}
Sergi Caelles, Jordi Pont-Tuset, Federico Perazzi, Alberto Montes, Kevis-Kokitsi Maninis, and Luc Van~Gool.
\newblock The 2019 davis challenge on vos: Unsupervised multi-object segmentation.
\newblock \emph{arXiv preprint arXiv:1905.00737}, 2019.

\bibitem[Chang et~al.(2021)Chang, Trafton, McCurry, and Thomaz]{chang2021unfair}
Mai~Lee Chang, Greg Trafton, J~Malcolm McCurry, and Andrea~Lockerd Thomaz.
\newblock Unfair! perceptions of fairness in human-robot teams.
\newblock In \emph{30th IEEE International Conference on Robot \& Human Interactive Communication (RO-MAN)}, pages 905--912, 2021.

\bibitem[Cheng et~al.(2020)Cheng, Collins, Zhu, Liu, Huang, Adam, and Chen]{cheng2020panoptic}
Bowen Cheng, Maxwell~D Collins, Yukun Zhu, Ting Liu, Thomas~S Huang, Hartwig Adam, and Liang-Chieh Chen.
\newblock Panoptic-deeplab: A simple, strong, and fast baseline for bottom-up panoptic segmentation.
\newblock In \emph{Proceedings of the IEEE/CVF conference on computer vision and pattern recognition}, pages 12475--12485, 2020.

\bibitem[Cheng et~al.(2022)Cheng, Misra, Schwing, Kirillov, and Girdhar]{cheng2022masked}
Bowen Cheng, Ishan Misra, Alexander~G Schwing, Alexander Kirillov, and Rohit Girdhar.
\newblock Masked-attention mask transformer for universal image segmentation.
\newblock In \emph{Proceedings of the IEEE/CVF conference on computer vision and pattern recognition}, pages 1290--1299, 2022.

\bibitem[Cheng and Schwing(2022)]{cheng2022xmem}
Ho~Kei Cheng and Alexander~G Schwing.
\newblock {XMem}: Long-term video object segmentation with an atkinson-shiffrin memory model.
\newblock In \emph{European Conference on Computer Vision}, pages 640--658. Springer, 2022.

\bibitem[Cho et~al.(2021)Cho, Hong, Jeon, Lee, Sohn, and Kim]{cho2021cats}
Seokju Cho, Sunghwan Hong, Sangryul Jeon, Yunsung Lee, Kwanghoon Sohn, and Seungryong Kim.
\newblock Cats: Cost aggregation transformers for visual correspondence.
\newblock \emph{Advances in Neural Information Processing Systems}, 34:\penalty0 9011--9023, 2021.

\bibitem[Ciaparrone et~al.(2020)Ciaparrone, S{\'a}nchez, Tabik, Troiano, Tagliaferri, and Herrera]{ciaparrone2020deep}
Gioele Ciaparrone, Francisco~Luque S{\'a}nchez, Siham Tabik, Luigi Troiano, Roberto Tagliaferri, and Francisco Herrera.
\newblock Deep learning in video multi-object tracking: A survey.
\newblock \emph{Neurocomputing}, 381:\penalty0 61--88, 2020.

\bibitem[Claure et~al.(2022)Claure, Chang, Kim, Omeiza, Brandao, Lee, and Jung]{claure2022fairness}
Houston Claure, Mai~Lee Chang, Seyun Kim, Daniel Omeiza, Martim Brandao, Min~Kyung Lee, and Malte Jung.
\newblock Fairness and transparency in human-robot interaction.
\newblock In \emph{17th ACM/IEEE International Conference on Human-Robot Interaction (HRI)}, pages 1244--1246, 2022.

\bibitem[Damen et~al.(2018)Damen, Doughty, Farinella, Fidler, Furnari, Kazakos, Moltisanti, Munro, Perrett, Price, et~al.]{damen2018scaling}
Dima Damen, Hazel Doughty, Giovanni~Maria Farinella, Sanja Fidler, Antonino Furnari, Evangelos Kazakos, Davide Moltisanti, Jonathan Munro, Toby Perrett, Will Price, et~al.
\newblock Scaling egocentric vision: The {Epic-Kitchens} dataset.
\newblock In \emph{Proceedings of the European conference on computer vision (ECCV)}, pages 720--736, 2018.

\bibitem[DeTone et~al.(2018)DeTone, Malisiewicz, and Rabinovich]{detone2018superpoint}
Daniel DeTone, Tomasz Malisiewicz, and Andrew Rabinovich.
\newblock {SuperPoint}: Self-supervised interest point detection and description.
\newblock In \emph{Proceedings of the IEEE conference on computer vision and pattern recognition workshops}, pages 224--236, 2018.

\bibitem[Donahue and Elhamifar(2024)]{donahue2024learning}
Gerard Donahue and Ehsan Elhamifar.
\newblock Learning to predict activity progress by self-supervised video alignment.
\newblock In \emph{Proceedings of the IEEE/CVF Conference on Computer Vision and Pattern Recognition}, pages 18667--18677, 2024.

\bibitem[Dou et~al.(2024)Dou, Yang, Nagarajan, Wang, Huang, Peng, Kitani, and Chu]{dou2024unlocking}
Zi-Yi Dou, Xitong Yang, Tushar Nagarajan, Huiyu Wang, Jing Huang, Nanyun Peng, Kris Kitani, and Fu-Jen Chu.
\newblock Unlocking exocentric video-language data for egocentric video representation learning.
\newblock \emph{arXiv preprint arXiv:2408.03567}, 2024.

\bibitem[Edstedt et~al.(2024)Edstedt, Sun, B{\"o}kman, Wadenb{\"a}ck, and Felsberg]{edstedt2024roma}
Johan Edstedt, Qiyu Sun, Georg B{\"o}kman, M{\aa}rten Wadenb{\"a}ck, and Michael Felsberg.
\newblock Roma: Robust dense feature matching.
\newblock In \emph{Proceedings of the IEEE/CVF Conference on Computer Vision and Pattern Recognition}, pages 19790--19800, 2024.

\bibitem[Fu et~al.(2024)Fu, Wang, Fu, Paudel, Huang, and Van~Gool]{fu2024objectrelator}
Yuqian Fu, Runze Wang, Yanwei Fu, Danda~Pani Paudel, Xuanjing Huang, and Luc Van~Gool.
\newblock Objectrelator: Enabling cross-view object relation understanding in ego-centric and exo-centric videos.
\newblock \emph{arXiv preprint arXiv:2411.19083}, 2024.

\bibitem[Furnari and Farinella(2020)]{furnari2020rolling}
Antonino Furnari and Giovanni~Maria Farinella.
\newblock Rolling-unrolling lstms for action anticipation from first-person video.
\newblock \emph{IEEE transactions on pattern analysis and machine intelligence}, 43\penalty0 (11):\penalty0 4021--4036, 2020.

\bibitem[Garg et~al.(2024)Garg, Puligilla, Kolathaya, Krishna, and Garg]{garg2024revisit}
Kartik Garg, Sai~Shubodh Puligilla, Shishir Kolathaya, Madhava Krishna, and Sourav Garg.
\newblock Revisit anything: Visual place recognition via image segment retrieval.
\newblock In \emph{European Conference on Computer Vision}, pages 326--343. Springer, 2024.

\bibitem[Goyal et~al.(2023)Goyal, Fan, Siam, and Sigal]{goyal2023tam}
Raghav Goyal, Wan-Cyuan Fan, Mennatullah Siam, and Leonid Sigal.
\newblock Tam-vt: Transformation-aware multi-scale video transformer for segmentation and tracking.
\newblock \emph{arXiv preprint arXiv:2312.08514}, 2023.

\bibitem[Grauman et~al.(2022)Grauman, Westbury, Byrne, Chavis, Furnari, Girdhar, Hamburger, Jiang, Liu, Liu, et~al.]{grauman2022ego4d}
Kristen Grauman, Andrew Westbury, Eugene Byrne, Zachary Chavis, Antonino Furnari, Rohit Girdhar, Jackson Hamburger, Hao Jiang, Miao Liu, Xingyu Liu, et~al.
\newblock Ego4d: Around the world in 3,000 hours of egocentric video.
\newblock In \emph{Proceedings of the IEEE/CVF Conference on Computer Vision and Pattern Recognition}, pages 18995--19012, 2022.

\bibitem[Grauman et~al.(2024)Grauman, Westbury, Torresani, Kitani, Malik, Afouras, Ashutosh, Baiyya, Bansal, Boote, et~al.]{grauman2024ego}
Kristen Grauman, Andrew Westbury, Lorenzo Torresani, Kris Kitani, Jitendra Malik, Triantafyllos Afouras, Kumar Ashutosh, Vijay Baiyya, Siddhant Bansal, Bikram Boote, et~al.
\newblock Ego-exo4d: Understanding skilled human activity from first-and third-person perspectives.
\newblock In \emph{Proceedings of the IEEE/CVF Conference on Computer Vision and Pattern Recognition}, pages 19383--19400, 2024.

\bibitem[He et~al.(2017)He, Gkioxari, Doll{\'a}r, and Girshick]{he2017mask}
Kaiming He, Georgia Gkioxari, Piotr Doll{\'a}r, and Ross Girshick.
\newblock Mask {R-CNN}.
\newblock In \emph{Proceedings of the IEEE international conference on computer vision}, pages 2961--2969, 2017.

\bibitem[Hutchinson and Gadepally(2021)]{hutchinson2021video}
Matthew~S Hutchinson and Vijay~N Gadepally.
\newblock Video action understanding.
\newblock \emph{IEEE Access}, 9:\penalty0 134611--134637, 2021.

\bibitem[Intraub(1997)]{intraub1997representation}
Helene Intraub.
\newblock The representation of visual scenes.
\newblock \emph{Trends in cognitive sciences}, 1\penalty0 (6):\penalty0 217--222, 1997.

\bibitem[Jiang et~al.(2021)Jiang, Trulls, Hosang, Tagliasacchi, and Yi]{jiang2021cotr}
Wei Jiang, Eduard Trulls, Jan Hosang, Andrea Tagliasacchi, and Kwang~Moo Yi.
\newblock Cotr: Correspondence transformer for matching across images.
\newblock In \emph{Proceedings of the IEEE/CVF International Conference on Computer Vision}, pages 6207--6217, 2021.

\bibitem[Karrer et~al.(2018)Karrer, Agarwal, Kamel, Siegwart, and Chli]{karrer2018collaborative}
Marco Karrer, Mohit Agarwal, Mina Kamel, Roland Siegwart, and Margarita Chli.
\newblock Collaborative 6dof relative pose estimation for two uavs with overlapping fields of view.
\newblock In \emph{IEEE International Conference on Robotics and Automation (ICRA)}, pages 6688--6693, 2018.

\bibitem[Ke et~al.(2023)Ke, Ye, Danelljan, Tai, Tang, Yu, et~al.]{ke2023segment}
Lei Ke, Mingqiao Ye, Martin Danelljan, Yu-Wing Tai, Chi-Keung Tang, Fisher Yu, et~al.
\newblock Segment anything in high quality.
\newblock \emph{Advances in Neural Information Processing Systems}, 36:\penalty0 29914--29934, 2023.

\bibitem[Kim et~al.(2020)Kim, Woo, Lee, and Kweon]{kim2020video}
Dahun Kim, Sanghyun Woo, Joon-Young Lee, and In~So Kweon.
\newblock Video panoptic segmentation.
\newblock In \emph{Proceedings of the IEEE/CVF conference on computer vision and pattern recognition}, pages 9859--9868, 2020.

\bibitem[Kirillov et~al.(2019)Kirillov, He, Girshick, Rother, and Doll{\'a}r]{kirillov2019panoptic}
Alexander Kirillov, Kaiming He, Ross Girshick, Carsten Rother, and Piotr Doll{\'a}r.
\newblock Panoptic segmentation.
\newblock In \emph{Proceedings of the IEEE/CVF conference on computer vision and pattern recognition}, pages 9404--9413, 2019.

\bibitem[Kirillov et~al.(2023)Kirillov, Mintun, Ravi, Mao, Rolland, Gustafson, Xiao, Whitehead, Berg, Lo, et~al.]{kirillov2023segment}
Alexander Kirillov, Eric Mintun, Nikhila Ravi, Hanzi Mao, Chloe Rolland, Laura Gustafson, Tete Xiao, Spencer Whitehead, Alexander~C Berg, Wan-Yen Lo, et~al.
\newblock Segment anything.
\newblock In \emph{Proceedings of the IEEE/CVF international conference on computer vision}, pages 4015--4026, 2023.

\bibitem[Lai et~al.(2024)Lai, Tian, Chen, Li, Yuan, Liu, and Jia]{lai2024lisa}
Xin Lai, Zhuotao Tian, Yukang Chen, Yanwei Li, Yuhui Yuan, Shu Liu, and Jiaya Jia.
\newblock Lisa: Reasoning segmentation via large language model.
\newblock In \emph{Proceedings of the IEEE/CVF Conference on Computer Vision and Pattern Recognition}, pages 9579--9589, 2024.

\bibitem[Li et~al.(2023{\natexlab{a}})Li, Jampani, Sun, and Sevilla-Lara]{li2023locate}
Gen Li, Varun Jampani, Deqing Sun, and Laura Sevilla-Lara.
\newblock Locate: Localize and transfer object parts for weakly supervised affordance grounding.
\newblock In \emph{Proceedings of the IEEE/CVF Conference on Computer Vision and Pattern Recognition}, pages 10922--10931, 2023{\natexlab{a}}.

\bibitem[Li et~al.(2024{\natexlab{a}})Li, Li, Wang, He, Wang, Wang, and Qiao]{li2024videomamba}
Kunchang Li, Xinhao Li, Yi Wang, Yinan He, Yali Wang, Limin Wang, and Yu Qiao.
\newblock Videomamba: State space model for efficient video understanding.
\newblock In \emph{European Conference on Computer Vision}, pages 237--255. Springer, 2024{\natexlab{a}}.

\bibitem[Li et~al.(2023{\natexlab{b}})Li, Wang, Zhou, Li, and Yang]{li2023unified}
Liulei Li, Wenguan Wang, Tianfei Zhou, Jianwu Li, and Yi Yang.
\newblock Unified mask embedding and correspondence learning for self-supervised video segmentation.
\newblock In \emph{Proceedings of the IEEE/CVF Conference on Computer Vision and Pattern Recognition}, pages 18706--18716, 2023{\natexlab{b}}.

\bibitem[Li et~al.(2024{\natexlab{b}})Li, Yuan, Li, Ding, Wu, Zhang, Li, Chen, and Loy]{li2024omg}
Xiangtai Li, Haobo Yuan, Wei Li, Henghui Ding, Size Wu, Wenwei Zhang, Yining Li, Kai Chen, and Chen~Change Loy.
\newblock Omg-seg: Is one model good enough for all segmentation?
\newblock In \emph{Proceedings of the IEEE/CVF conference on computer vision and pattern recognition}, pages 27948--27959, 2024{\natexlab{b}}.

\bibitem[Li et~al.(2021)Li, Nagarajan, Xiong, and Grauman]{li2021ego}
Yanghao Li, Tushar Nagarajan, Bo Xiong, and Kristen Grauman.
\newblock Ego-exo: Transferring visual representations from third-person to first-person videos.
\newblock In \emph{Proceedings of the IEEE/CVF Conference on Computer Vision and Pattern Recognition}, pages 6943--6953, 2021.

\bibitem[Liang et~al.(2023)Liang, Wu, Dai, Li, Zhao, Zhang, Zhang, Vajda, and Marculescu]{liang2023open}
Feng Liang, Bichen Wu, Xiaoliang Dai, Kunpeng Li, Yinan Zhao, Hang Zhang, Peizhao Zhang, Peter Vajda, and Diana Marculescu.
\newblock Open-vocabulary semantic segmentation with mask-adapted clip.
\newblock In \emph{Proceedings of the IEEE/CVF conference on computer vision and pattern recognition}, pages 7061--7070, 2023.

\bibitem[Lindenberger et~al.(2023)Lindenberger, Sarlin, and Pollefeys]{lindenberger2023lightglue}
Philipp Lindenberger, Paul-Edouard Sarlin, and Marc Pollefeys.
\newblock Lightglue: Local feature matching at light speed.
\newblock In \emph{Proceedings of the IEEE/CVF International Conference on Computer Vision}, pages 17627--17638, 2023.

\bibitem[Liu et~al.(2020)Liu, Zhu, Yamada, and Yang]{liu2020semantic}
Yanbin Liu, Linchao Zhu, Makoto Yamada, and Yi Yang.
\newblock Semantic correspondence as an optimal transport problem.
\newblock In \emph{Proceedings of the IEEE/CVF conference on computer vision and pattern recognition}, pages 4463--4472, 2020.

\bibitem[Long et~al.(2015)Long, Shelhamer, and Darrell]{long2015fully}
Jonathan Long, Evan Shelhamer, and Trevor Darrell.
\newblock Fully convolutional networks for semantic segmentation.
\newblock In \emph{Proceedings of the IEEE conference on computer vision and pattern recognition}, pages 3431--3440, 2015.

\bibitem[Loshchilov and Hutter(2019)]{loshchilov2017decoupled}
Ilya Loshchilov and Frank Hutter.
\newblock Decoupled weight decay regularization.
\newblock In \emph{International Conference on Learning Representations, {ICLR}}, 2019.

\bibitem[Lowe(2004)]{lowe2004distinctive}
David~G Lowe.
\newblock Distinctive image features from scale-invariant keypoints.
\newblock \emph{International journal of computer vision}, 60:\penalty0 91--110, 2004.

\bibitem[Mai et~al.(2023)Mai, Hamdi, Giancola, Zhao, and Ghanem]{mai2023egoloc}
Jinjie Mai, Abdullah Hamdi, Silvio Giancola, Chen Zhao, and Bernard Ghanem.
\newblock Egoloc: Revisiting 3d object localization from egocentric videos with visual queries.
\newblock In \emph{Proceedings of the IEEE/CVF International Conference on Computer Vision}, pages 45--57, 2023.

\bibitem[Minderer et~al.(2023)Minderer, Gritsenko, and Houlsby]{minderer2023scaling}
Matthias Minderer, Alexey Gritsenko, and Neil Houlsby.
\newblock Scaling open-vocabulary object detection.
\newblock \emph{Advances in Neural Information Processing Systems}, 36:\penalty0 72983--73007, 2023.

\bibitem[Mohan and Valada(2021)]{mohan2021efficientps}
Rohit Mohan and Abhinav Valada.
\newblock Efficientps: Efficient panoptic segmentation.
\newblock \emph{International Journal of Computer Vision}, 129\penalty0 (5):\penalty0 1551--1579, 2021.

\bibitem[Mur-Labadia et~al.(2023{\natexlab{a}})Mur-Labadia, Guerrero, and Martinez-Cantin]{mur2023multi}
Lorenzo Mur-Labadia, Jose~J Guerrero, and Ruben Martinez-Cantin.
\newblock Multi-label affordance mapping from egocentric vision.
\newblock In \emph{Proceedings of the IEEE/CVF International Conference on Computer Vision}, pages 5238--5249, 2023{\natexlab{a}}.

\bibitem[Mur-Labadia et~al.(2023{\natexlab{b}})Mur-Labadia, Martinez-Cantin, and Guerrero]{mur2023bayesian}
Lorenzo Mur-Labadia, Ruben Martinez-Cantin, and Jose~J Guerrero.
\newblock Bayesian deep learning for affordance segmentation in images.
\newblock In \emph{IEEE International Conference on Robotics and Automation (ICRA)}, pages 6981--6987, 2023{\natexlab{b}}.

\bibitem[Mur-Labadia et~al.(2024)Mur-Labadia, Martinez-Cantin, Guerrero, Farinella, and Furnari]{mur2024aff}
Lorenzo Mur-Labadia, Ruben Martinez-Cantin, Jose~J Guerrero, Giovanni~Maria Farinella, and Antonino Furnari.
\newblock Aff-ttention! affordances and attention models for short-term object interaction anticipation.
\newblock In \emph{European Conference on Computer Vision}, pages 167--184. Springer, 2024.

\bibitem[Nagarajan et~al.(2020)Nagarajan, Li, Feichtenhofer, and Grauman]{nagarajan2020ego}
Tushar Nagarajan, Yanghao Li, Christoph Feichtenhofer, and Kristen Grauman.
\newblock Ego-topo: Environment affordances from egocentric video.
\newblock In \emph{Proceedings of the IEEE/CVF Conference on Computer Vision and Pattern Recognition}, pages 163--172, 2020.

\bibitem[Oh et~al.(2019)Oh, Lee, Xu, and Kim]{Oh_2019_ICCV}
Seoung~Wug Oh, Joon-Young Lee, Ning Xu, and Seon~Joo Kim.
\newblock Video object segmentation using space-time memory networks.
\newblock In \emph{Proceedings of the IEEE/CVF International Conference on Computer Vision (ICCV)}, 2019.

\bibitem[Oord et~al.(2018)Oord, Li, and Vinyals]{oord2018representation}
Aaron van~den Oord, Yazhe Li, and Oriol Vinyals.
\newblock Representation learning with contrastive predictive coding.
\newblock \emph{arXiv preprint arXiv:1807.03748}, 2018.

\bibitem[Oquab et~al.(2023)Oquab, Darcet, Moutakanni, Vo, Szafraniec, Khalidov, Fernandez, Haziza, Massa, El-Nouby, et~al.]{oquab2023dinov2}
Maxime Oquab, Timoth{\'e}e Darcet, Th{\'e}o Moutakanni, Huy Vo, Marc Szafraniec, Vasil Khalidov, Pierre Fernandez, Daniel Haziza, Francisco Massa, Alaaeldin El-Nouby, et~al.
\newblock {DINOv2}: Learning robust visual features without supervision.
\newblock \emph{arXiv preprint arXiv:2304.07193}, 2023.

\bibitem[Perazzi et~al.(2016)Perazzi, Pont-Tuset, McWilliams, Van~Gool, Gross, and Sorkine-Hornung]{perazzi2016benchmark}
Federico Perazzi, Jordi Pont-Tuset, Brian McWilliams, Luc Van~Gool, Markus Gross, and Alexander Sorkine-Hornung.
\newblock A benchmark dataset and evaluation methodology for video object segmentation.
\newblock In \emph{Proceedings of the IEEE conference on computer vision and pattern recognition}, pages 724--732, 2016.

\bibitem[Quattrocchi et~al.(2024)Quattrocchi, Furnari, Di~Mauro, Giuffrida, and Farinella]{quattrocchi2024synchronization}
Camillo Quattrocchi, Antonino Furnari, Daniele Di~Mauro, Mario~Valerio Giuffrida, and Giovanni~Maria Farinella.
\newblock Synchronization is all you need: Exocentric-to-egocentric transfer for temporal action segmentation with unlabeled synchronized video pairs.
\newblock In \emph{European Conference on Computer Vision}, pages 253--270. Springer, 2024.

\bibitem[Radevski et~al.(2023)Radevski, Grujicic, Blaschko, Moens, and Tuytelaars]{radevski2023multimodal}
Gorjan Radevski, Dusan Grujicic, Matthew Blaschko, Marie-Francine Moens, and Tinne Tuytelaars.
\newblock Multimodal distillation for egocentric action recognition.
\newblock In \emph{Proceedings of the IEEE/CVF International Conference on Computer Vision}, pages 5213--5224, 2023.

\bibitem[Radford et~al.(2021)Radford, Kim, Hallacy, Ramesh, Goh, Agarwal, Sastry, Askell, Mishkin, Clark, et~al.]{radford2021learning}
Alec Radford, Jong~Wook Kim, Chris Hallacy, Aditya Ramesh, Gabriel Goh, Sandhini Agarwal, Girish Sastry, Amanda Askell, Pamela Mishkin, Jack Clark, et~al.
\newblock Learning transferable visual models from natural language supervision.
\newblock In \emph{International conference on machine learning}, pages 8748--8763. PmLR, 2021.

\bibitem[Rahmani and Mian(2016)]{rahmani2016knowledge}
H. Rahmani and A. Mian.
\newblock Learning a non-linear knowledge transfer model for cross-view action recognition.
\newblock In \emph{CVPR}, 2016.

\bibitem[Rubinstein et~al.(2013)Rubinstein, Joulin, Kopf, and Liu]{rubinstein2013unsupervised}
Michael Rubinstein, Armand Joulin, Johannes Kopf, and Ce Liu.
\newblock Unsupervised joint object discovery and segmentation in internet images.
\newblock In \emph{Proceedings of the IEEE conference on computer vision and pattern recognition}, pages 1939--1946, 2013.

\bibitem[Rublee et~al.(2011)Rublee, Rabaud, Konolige, and Bradski]{rublee2011orb}
Ethan Rublee, Vincent Rabaud, Kurt Konolige, and Gary Bradski.
\newblock {ORB}: An efficient alternative to {SIFT} or {SURF}.
\newblock In \emph{2011 International conference on computer vision}, pages 2564--2571. Ieee, 2011.

\bibitem[Russakovsky et~al.(2015)Russakovsky, Deng, Su, Krause, Satheesh, Ma, Huang, Karpathy, Khosla, Bernstein, et~al.]{russakovsky2015imagenet}
Olga Russakovsky, Jia Deng, Hao Su, Jonathan Krause, Sanjeev Satheesh, Sean Ma, Zhiheng Huang, Andrej Karpathy, Aditya Khosla, Michael Bernstein, et~al.
\newblock Imagenet large scale visual recognition challenge.
\newblock \emph{International journal of computer vision}, 115:\penalty0 211--252, 2015.

\bibitem[Sarlin et~al.(2020)Sarlin, DeTone, Malisiewicz, and Rabinovich]{sarlin2020superglue}
Paul-Edouard Sarlin, Daniel DeTone, Tomasz Malisiewicz, and Andrew Rabinovich.
\newblock Superglue: Learning feature matching with graph neural networks.
\newblock In \emph{Proceedings of the IEEE/CVF conference on computer vision and pattern recognition}, pages 4938--4947, 2020.

\bibitem[Schmuck and Chli(2019)]{schmuck2019ccm}
Patrik Schmuck and Margarita Chli.
\newblock Ccm-slam: Robust and efficient centralized collaborative monocular simultaneous localization and mapping for robotic teams.
\newblock \emph{Journal of Field Robotics}, 36\penalty0 (4):\penalty0 763--781, 2019.

\bibitem[Shen et~al.(2022)Shen, Efros, Joulin, and Aubry]{shen2022learning}
Xi Shen, Alexei~A Efros, Armand Joulin, and Mathieu Aubry.
\newblock Learning co-segmentation by segment swapping for retrieval and discovery.
\newblock In \emph{Proceedings of the IEEE/CVF Conference on Computer Vision and Pattern Recognition}, pages 5082--5092, 2022.

\bibitem[Shlapentokh-Rothman et~al.(2024)Shlapentokh-Rothman, Blume, Xiao, Wu, TV, Tao, Lee, Torres, Wang, and Hoiem]{shlapentokh2024region}
Michal Shlapentokh-Rothman, Ansel Blume, Yao Xiao, Yuqun Wu, Sethuraman TV, Heyi Tao, Jae~Yong Lee, Wilfredo Torres, Yu-Xiong Wang, and Derek Hoiem.
\newblock Region-based representations revisited.
\newblock In \emph{Proceedings of the IEEE/CVF Conference on Computer Vision and Pattern Recognition}, pages 17107--17116, 2024.

\bibitem[Sun et~al.(2021)Sun, Shen, Wang, Bao, and Zhou]{sun2021loftr}
Jiaming Sun, Zehong Shen, Yuang Wang, Hujun Bao, and Xiaowei Zhou.
\newblock Loftr: Detector-free local feature matching with transformers.
\newblock In \emph{Proceedings of the IEEE/CVF conference on computer vision and pattern recognition}, pages 8922--8931, 2021.

\bibitem[Taniai et~al.(2016)Taniai, Sinha, and Sato]{taniai2016joint}
Tatsunori Taniai, Sudipta~N Sinha, and Yoichi Sato.
\newblock Joint recovery of dense correspondence and cosegmentation in two images.
\newblock In \emph{Proceedings of the IEEE conference on computer vision and pattern recognition}, pages 4246--4255, 2016.

\bibitem[Tokmakov et~al.(2017)Tokmakov, Alahari, and Schmid]{tokmakov2017learning}
Pavel Tokmakov, Karteek Alahari, and Cordelia Schmid.
\newblock Learning video object segmentation with visual memory.
\newblock In \emph{Proceedings of the IEEE international conference on computer vision}, pages 4481--4490, 2017.

\bibitem[Tran et~al.(2023)Tran, Brown, Weidlich, Billinghurst, and Parker]{tran2023wearable}
Tram Thi~Minh Tran, Shane Brown, Oliver Weidlich, Mark Billinghurst, and Callum Parker.
\newblock Wearable augmented reality: Research trends and future directions from three major venues.
\newblock \emph{IEEE Transactions on Visualization and Computer Graphics}, 29\penalty0 (11):\penalty0 4782--4793, 2023.

\bibitem[Tsai et~al.(2016)Tsai, Yang, and Black]{tsai2016video}
Yi-Hsuan Tsai, Ming-Hsuan Yang, and Michael~J Black.
\newblock Video segmentation via object flow.
\newblock In \emph{Proceedings of the IEEE conference on computer vision and pattern recognition}, pages 3899--3908, 2016.

\bibitem[Tyszkiewicz et~al.(2020)Tyszkiewicz, Fua, and Trulls]{tyszkiewicz2020disk}
Micha{\l} Tyszkiewicz, Pascal Fua, and Eduard Trulls.
\newblock Disk: Learning local features with policy gradient.
\newblock \emph{Advances in Neural Information Processing Systems}, 33:\penalty0 14254--14265, 2020.

\bibitem[Vaswani et~al.(2017)Vaswani, Shazeer, Parmar, Uszkoreit, Jones, Gomez, Kaiser, and Polosukhin]{vaswani2017attention}
Ashish Vaswani, Noam Shazeer, Niki Parmar, Jakob Uszkoreit, Llion Jones, Aidan~N Gomez, {\L}ukasz Kaiser, and Illia Polosukhin.
\newblock Attention is all you need.
\newblock \emph{Advances in neural information processing systems}, 30, 2017.

\bibitem[Vyas and Bhatt(2017)]{vyas2017augmented}
Daiwat~Amit Vyas and Dvijesh Bhatt.
\newblock Augmented reality (ar) applications: A survey on current trends, challenges, \& future scope.
\newblock \emph{International Journal of Advanced Research in Computer Science}, 8\penalty0 (5), 2017.

\bibitem[Wang et~al.(2024)Wang, Ma, Xin, Hou, Xing, Dai, Wang, and Liu]{wang2024visual}
Mengmeng Wang, Teli Ma, Shuo Xin, Xiaojun Hou, Jiazheng Xing, Guang Dai, Jingdong Wang, and Yong Liu.
\newblock Visual object tracking across diverse data modalities: A review.
\newblock \emph{arXiv preprint arXiv:2412.09991}, 2024.

\bibitem[Woo et~al.(2021)Woo, Kim, Lee, and Kweon]{woo2021learning}
Sanghyun Woo, Dahun Kim, Joon-Young Lee, and In~So Kweon.
\newblock Learning to associate every segment for video panoptic segmentation.
\newblock In \emph{Proceedings of the IEEE/CVF Conference on Computer Vision and Pattern Recognition}, pages 2705--2714, 2021.

\bibitem[Xie et~al.(2021)Xie, Wang, Yu, Anandkumar, Alvarez, and Luo]{xie2021segformer}
Enze Xie, Wenhai Wang, Zhiding Yu, Anima Anandkumar, Jose~M Alvarez, and Ping Luo.
\newblock Segformer: Simple and efficient design for semantic segmentation with transformers.
\newblock \emph{Advances in neural information processing systems}, 34:\penalty0 12077--12090, 2021.

\bibitem[Xiong et~al.(2024)Xiong, Varadarajan, Wu, Xiang, Xiao, Zhu, Dai, Wang, Sun, Iandola, et~al.]{xiong2024efficientsam}
Yunyang Xiong, Bala Varadarajan, Lemeng Wu, Xiaoyu Xiang, Fanyi Xiao, Chenchen Zhu, Xiaoliang Dai, Dilin Wang, Fei Sun, Forrest Iandola, et~al.
\newblock Efficientsam: Leveraged masked image pretraining for efficient segment anything.
\newblock In \emph{Proceedings of the IEEE/CVF Conference on Computer Vision and Pattern Recognition}, pages 16111--16121, 2024.

\bibitem[Xu et~al.(2018)Xu, Yang, Fan, Yang, Yue, Liang, Price, Cohen, and Huang]{xu2018youtube}
Ning Xu, Linjie Yang, Yuchen Fan, Jianchao Yang, Dingcheng Yue, Yuchen Liang, Brian Price, Scott Cohen, and Thomas Huang.
\newblock Youtube-vos: Sequence-to-sequence video object segmentation.
\newblock In \emph{Proceedings of the European conference on computer vision (ECCV)}, pages 585--601, 2018.

\bibitem[Xue and Grauman(2023)]{xue2023learning}
Zihui~Sherry Xue and Kristen Grauman.
\newblock Learning fine-grained view-invariant representations from unpaired ego-exo videos via temporal alignment.
\newblock \emph{Advances in Neural Information Processing Systems}, 36:\penalty0 53688--53710, 2023.

\bibitem[Yang et~al.(2019)Yang, Wang, Bertinetto, Hu, Bai, and Torr]{yang2019anchor}
Zhao Yang, Qiang Wang, Luca Bertinetto, Weiming Hu, Song Bai, and Philip~HS Torr.
\newblock Anchor diffusion for unsupervised video object segmentation.
\newblock In \emph{Proceedings of the IEEE/CVF international conference on computer vision}, pages 931--940, 2019.

\bibitem[Yi et~al.(2016)Yi, Trulls, Lepetit, and Fua]{yi2016lift}
Kwang~Moo Yi, Eduard Trulls, Vincent Lepetit, and Pascal Fua.
\newblock Lift: Learned invariant feature transform.
\newblock In \emph{Computer Vision--ECCV 2016: 14th European Conference, Amsterdam, The Netherlands, October 11-14, 2016, Proceedings, Part VI 14}, pages 467--483. Springer, 2016.

\bibitem[Yu et~al.(2019)Yu, Cai, Liu, and Lu]{yu2019joint}
H. Yu, M. Cai, Y. Liu, and F. Lu.
\newblock Joint attention learning for first and third person video co-analysis.
\newblock \emph{ACM MM}, 2019.

\bibitem[Zhang et~al.(2023{\natexlab{a}})Zhang, Herrmann, Hur, Polania~Cabrera, Jampani, Sun, and Yang]{zhang2023tale}
Junyi Zhang, Charles Herrmann, Junhwa Hur, Luisa Polania~Cabrera, Varun Jampani, Deqing Sun, and Ming-Hsuan Yang.
\newblock A tale of two features: Stable diffusion complements dino for zero-shot semantic correspondence.
\newblock \emph{Advances in Neural Information Processing Systems}, 36:\penalty0 45533--45547, 2023{\natexlab{a}}.

\bibitem[Zhang et~al.(2023{\natexlab{b}})Zhang, Liu, Yang, Hu, Liu, and Stiefelhagen]{zhang2023cmx}
Jiaming Zhang, Huayao Liu, Kailun Yang, Xinxin Hu, Ruiping Liu, and Rainer Stiefelhagen.
\newblock Cmx: Cross-modal fusion for rgb-x semantic segmentation with transformers.
\newblock \emph{IEEE Transactions on intelligent transportation systems}, 24\penalty0 (12):\penalty0 14679--14694, 2023{\natexlab{b}}.

\bibitem[Zhang et~al.(2024)Zhang, Ma, Zhang, and Bai]{zhang2024psalm}
Zheng Zhang, Yeyao Ma, Enming Zhang, and Xiang Bai.
\newblock Psalm: Pixelwise segmentation with large multi-modal model.
\newblock In \emph{European Conference on Computer Vision}, pages 74--91. Springer, 2024.

\bibitem[Zhao et~al.(2023{\natexlab{a}})Zhao, Ding, An, Du, Yu, Li, Tang, and Wang]{zhao2023fast}
Xu Zhao, Wenchao Ding, Yongqi An, Yinglong Du, Tao Yu, Min Li, Ming Tang, and Jinqiao Wang.
\newblock Fast segment anything.
\newblock \emph{arXiv preprint arXiv:2306.12156}, 2023{\natexlab{a}}.

\bibitem[Zhao et~al.(2023{\natexlab{b}})Zhao, Wu, Chen, Chen, Xu, and Li]{zhao2023aliked}
Xiaoming Zhao, Xingming Wu, Weihai Chen, Peter~CY Chen, Qingsong Xu, and Zhengguo Li.
\newblock Aliked: A lighter keypoint and descriptor extraction network via deformable transformation.
\newblock \emph{IEEE Transactions on Instrumentation and Measurement}, 72:\penalty0 1--16, 2023{\natexlab{b}}.

\bibitem[Zou et~al.(2023)Zou, Dou, Yang, Gan, Li, Li, Dai, Behl, Wang, Yuan, et~al.]{zou2023generalized}
Xueyan Zou, Zi-Yi Dou, Jianwei Yang, Zhe Gan, Linjie Li, Chunyuan Li, Xiyang Dai, Harkirat Behl, Jianfeng Wang, Lu Yuan, et~al.
\newblock Generalized decoding for pixel, image, and language.
\newblock In \emph{Proceedings of the IEEE/CVF conference on computer vision and pattern recognition}, pages 15116--15127, 2023.

\end{thebibliography}
}

\clearpage 

\twocolumn[%
\begin{center}
    {\LARGE \bfseries Supplementary Material \par}
    {\large \itshape O-MaMa: Learning Object Mask Matching between Egocentric and Exocentric Views\par}
    \vspace{1.5em}
\end{center}]

\begin{appendices}
\pagenumbering{roman}
    \counterwithin{figure}{section}
    \counterwithin{table}{section}
    \counterwithin{equation}{section}
    
\cref{sec:cmx_implementation} of this supplementary material explains how CMX \cite{zhang2023cmx} was modified and retrained for our task. On the other hand, \cref{sec:geometry} explains in detail the geometry baselines.
We also report extra qualitative examples and attention maps for a more detailed comparison in \cref{sec:qualit}. 
    \section{Implementation details of CMX} \label{sec:cmx_implementation}

The original CMX \cite{zhang2023cmx} had two input images of three channels, we adapted it to input the source RGB channels concatenated with the object mask, and the destination RGB image. Since there are few examples of destination images without mask, we augment it with a probability of 10\%.
We use the version of \cite{zhang2023cmx} that incorporates  Mix Transformer encoder (MiT) \cite{xie2021segformer} pretrained on ImageNet \cite{russakovsky2015imagenet}. In particular, we use the variant MiT-b4, changing the patch embedding input size for the source input to 4 channels.
As for the decoder, we fuse the multi-level
features from the backbone as done originally, and input the fused features to the XSegTx \cite{grauman2024ego} decoder to generate
a mask prediction. During training, we found out that freezing the pre-trained weights and training the rest worked the best, whereas fine-tuning it afterwards or training without freezing layers led to no improvement or convergence. To prevent overfitting and speeding up the epoch training time, we randomly sampled 2\% of the training and validation set being different for each epoch over 25 epochs.
We employ AdamW optimizer \cite{loshchilov2017decoupled} with weight decay 0.01 and a starting learning rate of $1e^{-3}$, which is decreased towards 0 using a cosine scheduler.
    \begin{figure*}[t]
    \centering
    \includegraphics[width=0.99\linewidth]{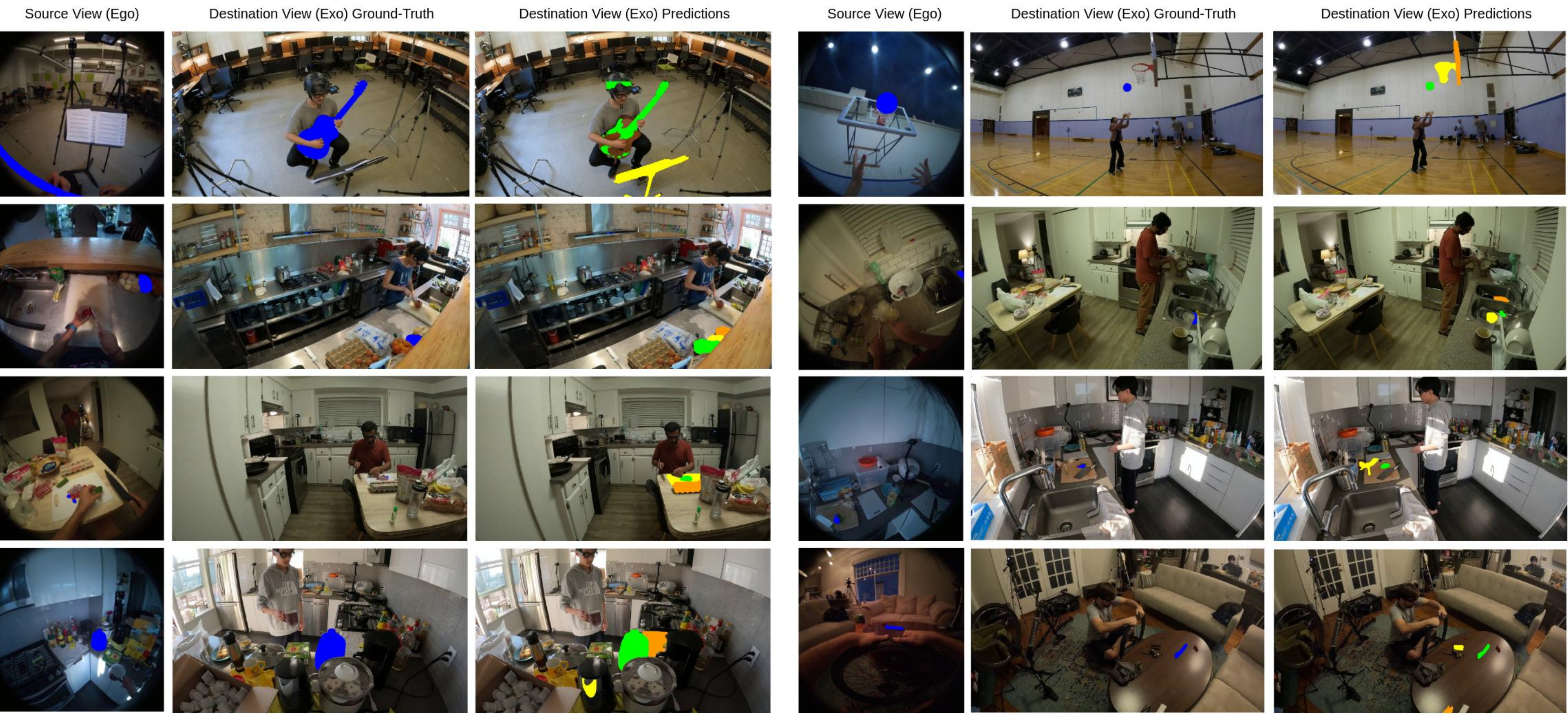}
    \caption{\textbf{Ego2Exo Extra Qualitative Results.} For visualization purposes, we show the top 3 masks in \textcolor{green}{green}, \textcolor{yellow}{yellow} and \textcolor{orange}{orange}. In \textcolor{blue}{blue} we show the source and ground-truth masks.}
    \label{fig:qualit1_supp}
\end{figure*}

\begin{figure*}[t]
    \centering
    \includegraphics[width=0.99\linewidth]{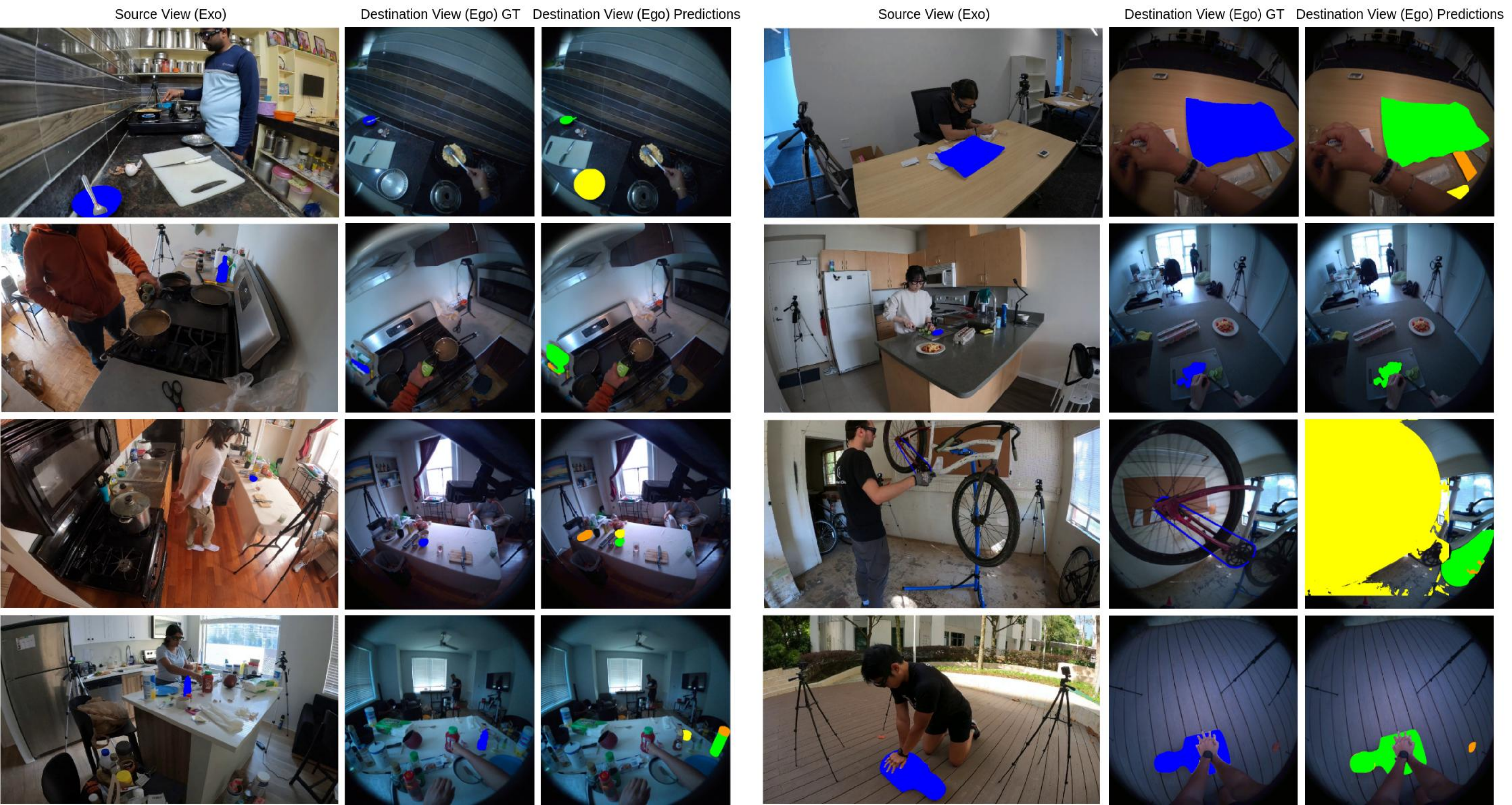}
    \caption{\textbf{Exo2Ego Extra Qualitative Results.} For visualization purposes, we show the top 3 masks in \textcolor{green}{green}, \textcolor{yellow}{yellow} and \textcolor{orange}{orange}. In \textcolor{blue}{blue} we show the source masks.}
    \label{fig:qualit2_supp}
\end{figure*}

\begin{figure*}[t]
    \centering
    \includegraphics[width=0.99\linewidth]{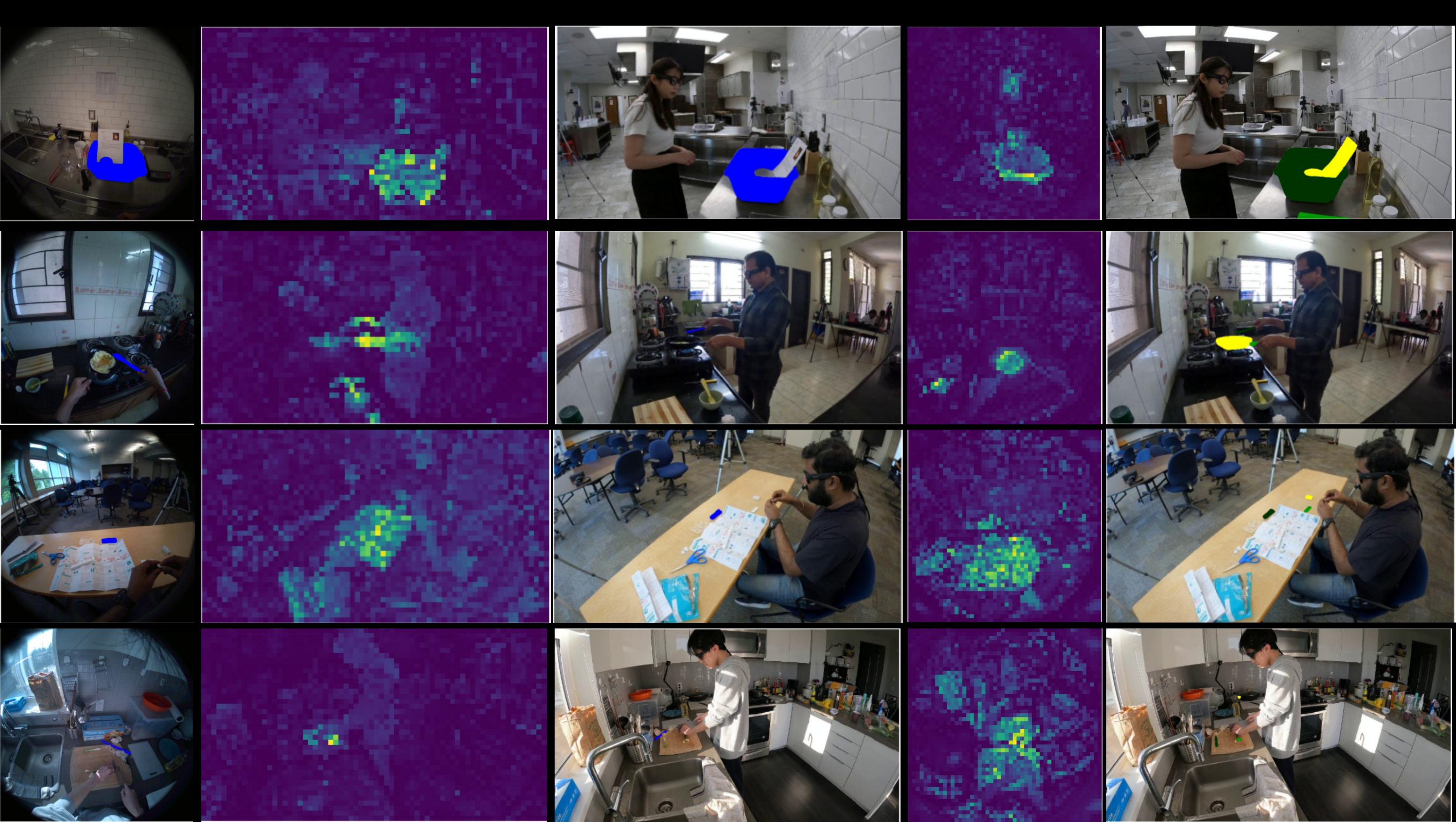}
    \caption{\textbf{Attention maps of the Ego$\leftrightarrow$Exo Cross Attention module.} We visualize the average of the attention maps, showing how the mechanism correlates the object features from the other image perspective.}
    \label{fig:qualit3_supp}
\end{figure*}

\section{Geometry Methods} \label{sec:geometry}

Since the k-NN baseline does not take into account the mask location, objects having similar appearance can lead to false positives. For that, we decided to restrict the previous approach to fulfill the epipolar line restriction by assuming a pin-hole camera model in a self-calibration fashion. Since our epipolar constraint has been defined with a wide threshold, this assumption is good enough to detect most false positives. For this end, we used RoMa~\citep{edstedt2024roma} to obtain the fundamental matrix $F$.
 
 Given the centroid of the source mask in homogeneous coordinates $\mathbf{x}^S = (x^S_x,x^S_y,1)$, we obtain the epipolar line in the destination image as $\mathbf{l}^D = F \cdot \mathbf{x}^S = (a,b,c)$. Then, we compute the perpendicular distance $d$ of the most feature-similar possible masks from the centroid to its corresponding epipolar line:

\begin{equation}
    d = \frac{|a \cdot x^S_{x} + b \cdot x^S_{y} + c|}{\sqrt{a^2 + b^2}}
\end{equation}

If the distance is superior to a certain threshold, the candidate mask is discarded.

\subsection{Success rate of matching methods}

The success rate we use to choose the features matching method is based on an epipolar geometry criterion. With the ground truth pose and the calibration of the cameras, we verify that the matches satisfy the epipolar constraint. Even when the pose estimation is not accurate enough to obtain a precise classification of the rate of correct matches in each pair of images, the obtained result is good enough for identifying success and failure cases.

\section{Extra Qualitative Results} \label{sec:qualit}

We report extra qualitative results in \cref{fig:qualit1_supp} and \cref{fig:qualit2_supp}, showing the best three FastSAM mask candidates predicted by our model for visualization purposes. Note that we only use the top-1 mask for reporting the official quantitative metrics. 
In most of the cases, the predicted mask matches with the target object, achieving a very fine-grained segmentation quality even when the source or target objects are considerably small.
However, our model is also dependent on the quality of the candidate masks, showing in some cases sparse segmentations (\ie the \textit{guitar} in \cref{fig:qualit1_supp}) or inaccurate masks (the \textit{chain} in \cref{fig:qualit2_supp}).

Finally, we visualize extra examples of the attention maps produced by our novel Ego$\leftrightarrow$Exo Cross Attention mechanism, which captures the object visual cues in the other viewpoint (\cref{fig:qualit3_supp}).

\end{appendices}

\end{document}